%% file: acl_latex.tex
\pdfoutput=1
\documentclass[11pt]{article}

\usepackage{acl}

\usepackage{times}
\usepackage{latexsym}

\usepackage[T1]{fontenc}
\usepackage[utf8]{inputenc}

\usepackage{microtype}

\usepackage{booktabs}
\usepackage{adjustbox}
\usepackage{caption}
\usepackage{multirow}
\usepackage{xcolor}
\usepackage{siunitx}

\usepackage{graphicx}
\usepackage{subcaption}

\usepackage{textcomp}
\usepackage{tabularx}

\usepackage[table]{xcolor} % 放在导言区
\usepackage{multirow}
\usepackage{amssymb} % for \checkmark
\usepackage{makecell}
\usepackage{enumitem}

\usepackage{array}
\usepackage{arydshln}

\definecolor{darkred}{HTML}{C00000}
\definecolor{darkblue}{HTML}{002060}

\usepackage{xcolor}
\usepackage{pifont} % 或 bbding/fontawesome5，按上面选一个
\newcommand{\equal}{\textsuperscript{\dag}}   % 共同一作 †
\newcommand{\corr}{\textsuperscript{*}}       % 通讯作者 *

\title{When Seeing Is not Enough: Revealing the Limits of Active Reasoning in MLLMs}

\author{Hongcheng Liu\equal, Pingjie Wang\equal, Yuhao Wang, Siqu Ou\\ \textbf{Yanfeng Wang, Yu Wang\corr} \\
Shanghai Jiao Tong University \\ \texttt{\{hongcheng\_liu,pingjiewang,colane,crataletio\}@sjtu.edu.cn} \\ \texttt{\{wangyanfeng, yuwangsjtu\}@sjtu.edu.cn} \\}

\begin{document}
\maketitle
\begingroup
\renewcommand{\thefootnote}{\fnsymbol{footnote}}
\footnotetext[2]{Equal contributions.}
\footnotetext[1]{Corresponding authors.}
\endgroup
\input{page/00Abstract}
\input{page/01Introduction}
\input{page/03Methods}
\input{page/04Experiments}
\input{page/05Analysis}
\input{page/06Conclusion}
\newpage
\newpage
\section*{Limitations}
Despite a two-stage procedure intended to improve responses and align scores with human judgments, the guess agent can still be inaccurate. Errors may stem from a distribution shift or propagation across stages. Moreover, although we carefully engineer prompts, a single prompt does not transfer reliably across models and can lead to mismatches at inference time. Overall performance may be marginally degraded. Furthermore, we set the maximum new tokens to 5000, which may be insufficient for some cases in thinking models and can lead to incomplete responses from MLLMs.

\section*{Ethical Considerations}
All images used in this study are sourced from the publicly available intrinsic images in the IIW dataset or generated using the publicly released Omnigen2. We include only non-sensitive content and exclude material that could reasonably be considered unsafe. All evaluated models are publicly available, and we used them strictly under their respective licenses and terms of use. Human ratings were provided by college students; no personal characteristics were requested, and no personally identifiable information was collected. Accordingly, we anticipate no specific ethical concerns arising from the data or evaluation procedures in this work. Furthermore,  we leveraged large language models as writing assistants for tasks such as rephrasing sentences, improving grammatical flow, and refining technical descriptions for clarity.

\newpage
\bibliography{anthology,custom}
\bibliographystyle{acl_natbib}
\clearpage
\appendix
\input{page/07Appendix}

\end{document}

%% file: page/00Abstract.tex
\begin{abstract}
Multimodal large language models (MLLMs) have shown strong capabilities across a broad range of benchmarks. However, most existing evaluations focus on passive inference, where models perform step-by-step reasoning under complete information. This setup is misaligned with real-world use, where seeing is not enough. This raises a fundamental question: \textbf{\textit{Can MLLMs actively acquire missing evidence under incomplete information?}} To bridge this gap, we require the MLLMs to actively acquire missing evidence and iteratively refine decisions under incomplete information, by selecting a target image from a candidate pool without task-specific priors. To support systematic study, we propose \textsc{GuessBench}, a benchmark with both perception-oriented and knowledge-oriented images for evaluating active reasoning in MLLMs. We evaluate 20 superior MLLMs and find that performance on active reasoning lags far behind it on passive settings, indicating substantial room for improvement. Further analysis identifies fine-grained perception and timely decision-making as key challenges. Ablation studies show that perceptual enhancements benefit smaller models, whereas thinking-oriented methods provide consistent gains across model sizes. These results suggest promising directions for future research on multimodal active reasoning.
\end{abstract}

% To bridge this gap, we frame the task as an interactive target guessing problem that requires selecting a target image from a candidate pool without task-specific priors. 

%% file: page/01Introduction.tex
\section{Introduction}
% Multimodal Large Language Models (MLLMs) have recently achieved state-of-the-art performance across a broad spectrum of vision–language benchmarks and real-world applications~\cite{ou2025bridgingdynamicperceptiongap,yin2024survey,li2025perception}. However, this success is validated predominantly under passive inference settings~\cite{wang2024mmsapcomprehensivebenchmarkassessing,liu2024mmbench}, where models are supplied with all necessary inputs to make a single-pass decision. This paradigm overlooks the fact that many real-world tasks require interaction and involve incomplete information~\cite{zhou2025passiveactivereasoninglarge}. For instance, an ideal intelligent agent, such as a product recommender, is expected to actively query user preferences and, through the presentation of images, efficiently narrow the solution space. This critical disparity raises a fundamental research question: \textbf{\textit{Can MLLMs actively acquire missing evidence under incomplete information scenarios?}}
Multimodal Large Language Models (MLLMs) have demonstrated impressive capabilities~\cite{ou2025bridgingdynamicperceptiongap,yin2024survey,li2025perception}, but their success is predominantly measured in passive inference settings~\cite{wang2024mmsapcomprehensivebenchmarkassessing,liu2024mmbench}, where they are supplied with all necessary information to make a single-pass decision. This benchmark-driven progress overlooks a fundamental reality: in the real world, seeing is often not enough~\cite{zhou2025passiveactivereasoninglarge}. Many tasks, from a product recommender learning user preferences to a robot navigating an unknown space, require interaction and reasoning under incomplete information~\cite{Raza2024ACR}. This critical disparity reveals a fundamental research question: \textbf{\textit{Can MLLMs actively acquire missing evidence when faced with uncertainty?}}

% a robot navigating an unknown space, require interaction and reasoning under incomplete information. 要加引用

\begin{figure}[t]
    \centering     
    \includegraphics[width=0.9\linewidth]{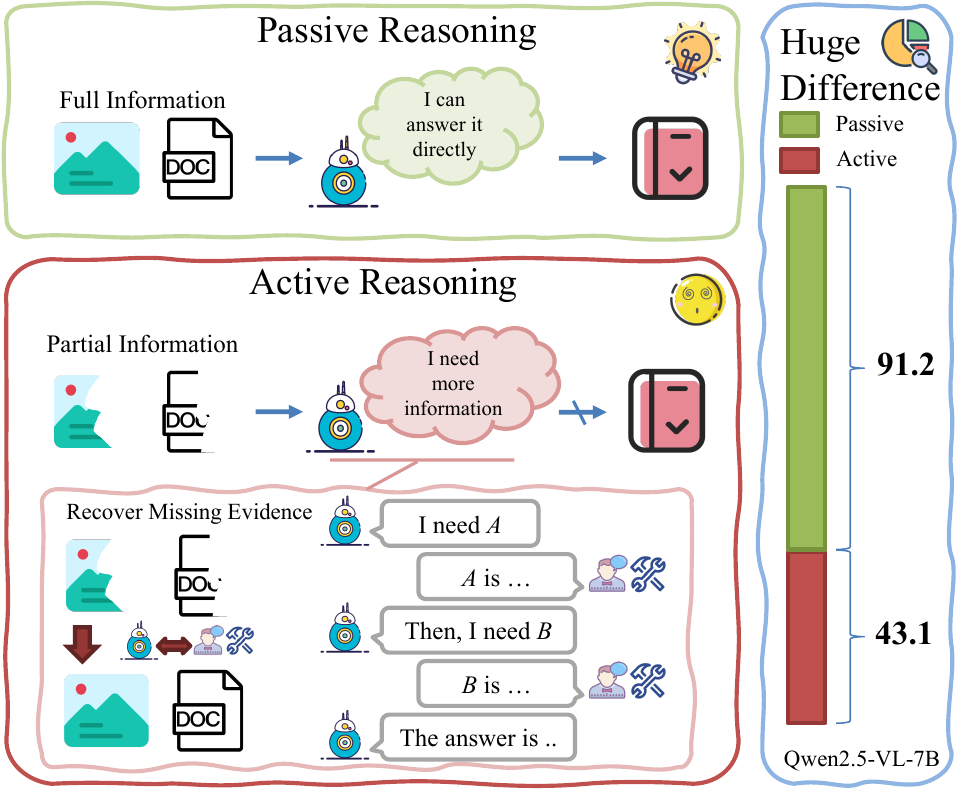}
    \caption{\textbf{Passive vs.\ active reasoning.} The left shows the difference between these two, and the right shows a pronounced gap between the two capabilities.
    }
    \label{fig:figure1}
    \vspace{-1.5em}
\end{figure}

% To bridge this gap, we introduce the active reasoning problem in multimodal fields, 
To bridge this gap, we introduce and formalize the active reasoning problem in multimodal contexts, as shown in Figure~\ref{fig:figure1}. Inspired by interactive games such as ``Guess Who I Am''~\cite{GuessWhoIAmApp} and ``GuessArena''~\cite{GuessArena}, we define this problem as an interactive target guessing problem, where an MLLM must select a target image from a candidate set without any auxiliary information at the start. To succeed, the model must compare candidates, ask strategic questions to acquire missing evidence, and decide whether to query further or commit to an answer. Success hinges on a goal-directed cognitive cycle that requires: (1) visual abstraction (\textbf{Perceive}), to capture commonalities and subtle distinctions within the pool; and (2) knowledge-integrated reasoning (\textbf{Think}), to combine visual cues, prior world knowledge, and responses from external sources into a strategic hypothesis.

Building on this paradigm, we propose \textsc{GuessBench}, the first systematic framework for rigorous evaluation of active reasoning in MLLMs. To enable comprehensive and fine-grained analysis, we construct a dataset that strategically mixes real-world images with nine distinct synthetic types. These types are organized into two categories: (1) perception-oriented categories, which emphasize the ability to capture subtle visual nuances such as facial details or minor structural variations. (2) knowledge-oriented categories, which require recalling external world knowledge such as specialized occupations or complex functional roles.

% Using this structured dataset, we rigorously benchmark 20 mainstream MLLMs, establishing a solid empirical foundation for this new paradigm. Our primary findings reveal several key conclusions.
% \begin{itemize}[itemsep=0pt, topsep=0pt, parsep=0pt]
%     \item \textbf{Challenge}: Models that perform well under passive evaluation degrade sharply on active tasks, with an average drop of \(35.2\%\).
%     \item \textbf{Perception is harder than knowledge}: Performance on perception-oriented images is \(27.5\%\) lower than knowledge-oriented images, indicating perception is more challenging.
%     \item \textbf{Scaling is not decisive}. MiMo-VL (7B) exceeds Qwen2.5-VL-32B under explicit thinking reinforcement learning, underscoring the impact of training strategy.
%     \item \textbf{Perception matters}: Varying fine-grained distinctions and candidate-pool size show that perception is pivotal; limited perceptual capacity constrains overall performance.
%     \item \textbf{Timely decisions are lacking}: Candidate-pool size versus progress reveals that models continue querying even when a single image already satisfies the query.
%     \item \textbf{Perception-enhanced methods aid small models}: Ablation experiments indicate that smaller models benefit substantially from improved perceptual capability.
%     \item \textbf{Explicit Thinking yields consistent improvement}: Both training- and prompt-based thinking modes improve performance by leveraging dynamic information more effectively.
% \end{itemize}
Using this structured dataset, we rigorously benchmark 20 mainstream MLLMs, establishing a solid empirical foundation for this new paradigm. We reveal that state-of-the-art models that excel under passive evaluation degrade markedly on active tasks, indicating substantial headroom for improvement. We further uncover two dominant impediments to active success: limited fine-grained perception and untimely decision-making. Our analysis also identifies clear pathways toward more powerful active reasoning, highlighting the benefits of perception-oriented enhancements for smaller models and the consistent gains from thinking-oriented methods across all models.

The main contributions can be summarized as:
\begin{itemize}[itemsep=0pt, topsep=0pt, parsep=0pt]
    \item \textbf{Multimodal active reasoning.} We identify and formalize the underexplored problem of multimodal active reasoning, where MLLMs must actively acquire missing evidence and iteratively refine decisions under incomplete information. This paradigm shifts the focus from passive inference to goal-directed, interactive decision cycles.
    \item \textbf{\textsc{GuessBench} framework.} We propose \textsc{GuessBench}, the first systematic framework for multimodal active reasoning. The dataset combines real-world images with nine distinct synthetic types, grouped into perception-oriented and knowledge-oriented categories, enabling comprehensive and fine-grained diagnostics of visual abstraction and knowledge-integrated reasoning.
    \item \textbf{Comprehensive analysis.} We empirically benchmark 20 mainstream MLLMs on \textsc{GuessBench} and find that they fall short in active reasoning scenarios. Further analyses reveal key limiting factors and point to effective enhancement strategies.
\end{itemize}

%% file: page/03Methods.tex
\section{GuessBench}
\begin{figure*}[t]
    \centering
    \includegraphics[width=.95\linewidth]{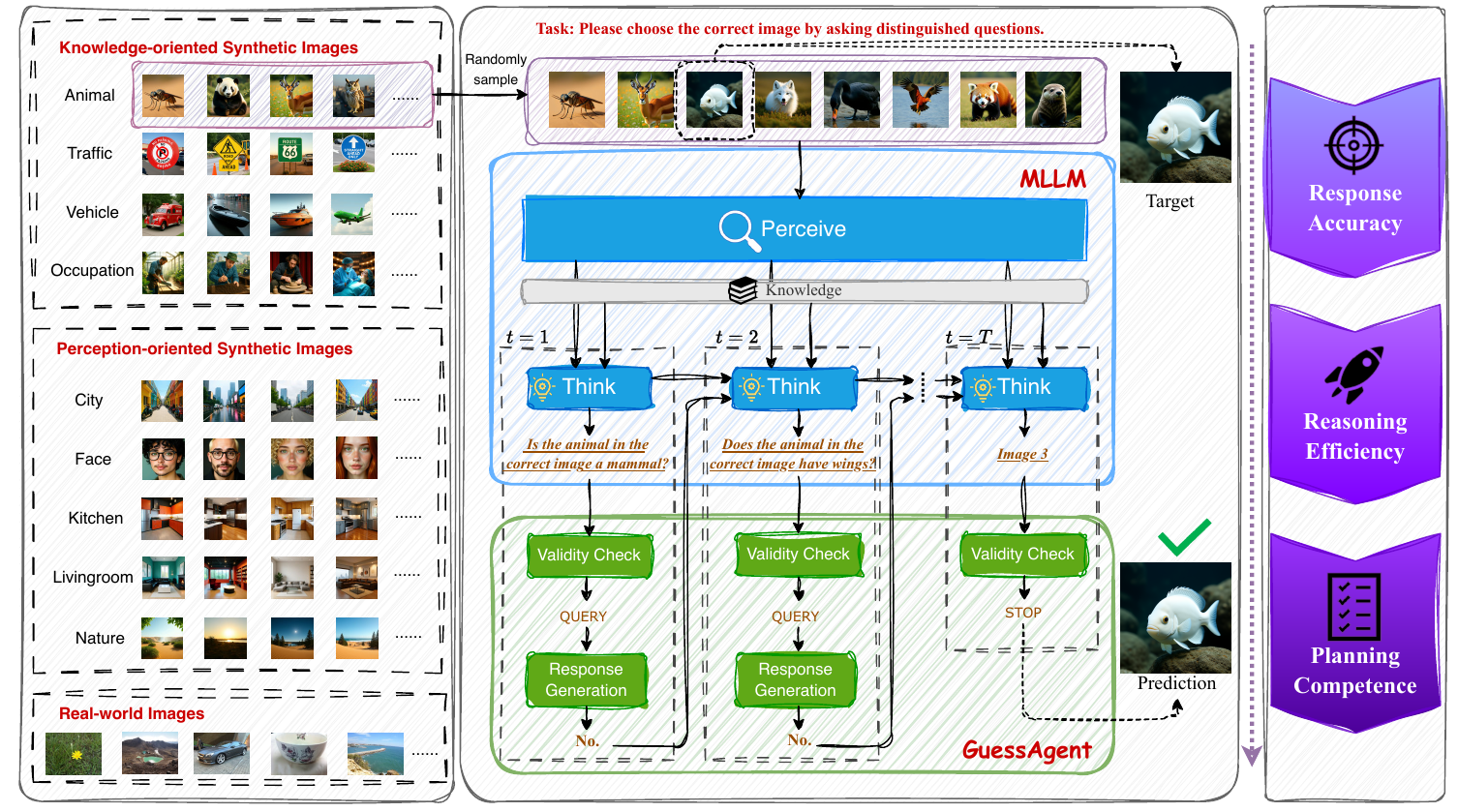}
    \caption{Overview of data distribution, interaction pipeline, and metric design.}
    \label{fig:overall}
    \vspace{-1.5em}
\end{figure*}
% To further assess the active reasoning capabilities of MLLMs, we propose a novel evaluation framework called GuessBench, which consists of both data construction and evaluation procedures. In this framework, the goal is to examine whether an MLLM can perform active reasoning under conditions of insufficient information by acquiring external evidence. Specifically, the model is required to identify a target image from a set of candidates based on limited reasoning cues, and it must actively obtain the missing evidence through interaction with external tools. The task is considered successful if the MLLM can locate the correct target image within a restricted reasoning steps.

\subsection{Evaluation Framework}
To push beyond passive evaluation, we introduce \textsc{GuessBench}, a systematic framework for rigorously testing the active reasoning capabilities of MLLMs under incomplete information. We formalize this multimodal challenge as an interactive target guessing problem: given a candidate pool and no prior clues, the model must select the single target image. Success requires an iterative perceive then think cycle. The model first (1) \textbf{Perceive} the visual candidates to extract shared and distinguishing features, then (2) \textbf{Think} by integrating these features with world knowledge to pose informative questions that acquire the missing evidence. This cycle repeats until the model commits to a confident final decision. The design of \textsc{GuessBench} is detailed in Figure~\ref{fig:overall} and the following sections.
% The design of GuessBench is detailed through its data construction ($\S\ref{subsec:data_construction}$), evaluation procedure ($\S\ref{subsec:evaluation_procedure}$), and evaluation metrics ($\S\ref{subsec:evaluation_metrics}$) in the sections that follow.
% 要考虑先说评估方式还是数据构造

\subsection{Data Construction}
\label{subsec:data_construction}
To comprehensively evaluate the active reasoning capabilities of MLLMs, we design a specialized dataset that integrates real-world and synthetic images, spanning nine domains for multi-faceted assessment. Real-world images are sampled from the IIW dataset~\cite{garg2024imageinwords}. For systematic tests of the core capabilities of perceiving and thinking, we engineer synthetic images in two categories: perception-oriented (city, face, kitchen, living room, nature) and knowledge-oriented (animal, traffic, vehicle, occupation). The former stresses sensitivity to subtle visual nuances among similar patterns, whereas the latter requires recalling external world knowledge to resolve complex distinctions.
For synthesis, we first use GPT-4o-mini~\cite{openai2024gpt4omini} to generate domain-specific, discriminative attributes, randomly compose them into full descriptions, and provide these descriptions to OmniGen2~\cite{wu2025omnigen2} to create images. We then employ Qwen2.5-VL-7B~\cite{bai2025qwen25vltechnicalreport} to produce detailed captions that extend the initial attributes and serve as a reference for external sources to answer the model’s strategic queries. Finally, we conduct human verification to rigorously validate the accuracy and quality of all synthesized images and captions.
% 要突出构造这三种数据的目的，为了测试模型的哪些方面的能力

\subsection{Evaluation Procedure}
\label{subsec:evaluation_procedure}

Inspired by the classic game ``Guess Who I Am?'', \textsc{GuessBench} formalizes the multimodal active reasoning problem as an interactive target guessing problem. Specifically, at the start of the evaluation session, $B$ images are randomly sampled from the data pool (detailed in \S\ref{subsec:data_construction}) to form an evaluation set $\mathcal{I}=\{ i^1, i^2, \dots, i^B\}$, in which a randomly selected image $i^\prime \in \mathcal{I}$ is designed as the target image.
Given the candidate pool $\mathcal{I}$, the MLLM ($\mathcal{M}$) must conduct an iterative query process by posing strategic questions ($q_t$) based on its own world-knowledge $\mathcal{K}_{\mathcal{M}}$ and obtaining facts ($a_t$) at each time step $t$. To keep queries concise and directly informative, the model is constrained to binary questions that admit Yes or No answers. This interactive process is repeated until the model determines its final prediction $\hat{i}$ which marks the end of the session at time step $T$. The entire decision-making process of the MLLM $\mathcal{M}$ at any step $t$ can be formally modeled as a function of the candidate pool $\mathcal{I}$, the history of previous questions and facts $\mathcal{H}_{t-1}=\{(q_1, a_1),\dots,(q_{t-1}, a_{t-1})\}$, and its internal knowledge $\mathcal{K}_{\mathcal{M}}$:
\begin{equation}
    <\delta_t, \text{Output}_t> = \mathcal{M}(\mathcal{I}, \mathcal{H}_{t-1}, \mathcal{K}_{\mathcal{M}}),
\end{equation}
where $\delta_t \in \{ \text{QUERY}, \text{STOP} \}$ is the decision type, and $\text{Output}_t$ is either the next question (if $\delta_t = \text{QUERY}$) or the final prediction $\hat{i}$ (if $\delta_t = \text{STOP}$).

To efficiently assist the MLLM in acquiring cues and to minimize incorrect or superfluous external responses, we design the GuessAgent to interpret and respond to the model's queries, which consists of a validity check and response generation components. For the validity check, we first employ Qwen3-8B~\cite{qwen3} to determine the decision type, classifying each model output into one of <QUERY>, <STOP>, or <INVALID>. A query is labeled <INVALID> if it cannot be answered with a simple Yes or No, thereby enforcing the binary-question constraint. For response generation, GuessAgent first attempts to retrieve relevant information from the image's detailed data attributes. If the question cannot be resolved using these attributes alone, then it accesses the comprehensive image captions and images. Details for GuessAgent are provided in the Appendix~\ref{app-prompt}.

\subsection{Evaluation Metrics}
\label{subsec:evaluation_metrics}
% To assess the active reasoning ability in both accuracy and efficiency across diverse domains, we propose a composite evaluation metric that integrates response accuracy ($A$), reasoning efficiency ($R$ ), and planning competence ($P$ ), drawing inspiration from GuessArena. The evaluation metric is as follows:

% \begin{equation}
% \text{metric} =  \text{A} * (\text{C} + R + P),
% \end{equation}
% where \text{C} is the constant to maintain the original weight of accuracy, and we set it to 1. In this way, the evaluation metric can be accomplished to primarily assess reasoning accuracy while also incorporating an assessment of reasoning efficiency and knowledge utilization. 

To comprehensively evaluate the active reasoning ability of MLLMs in terms of both accuracy and efficiency across diverse domains, we propose a composite evaluation metric $\mathcal{S}$ that integrates response accuracy ($\mathcal{A}$), reasoning efficiency ($\mathcal{R}$), and planning competence ($\mathcal{P}$). This metric draws inspiration from the design of GuessArena and is defined as follows:
\begin{equation}
\small
    \mathcal{S} = \mathcal{A}\cdot \frac{\omega + \mathcal{R} + \mathcal{P}}{\omega+2},
\end{equation}
where $\omega \geq 1$ is a constant that preserves the baseline contribution of accuracy, which is set as 1 by default. In this way, the metric can primarily assess identification accuracy while quantitatively rewarding efficiency and strategic knowledge utilization.

% \textbf{Response accuracy} evaluates whether the MLLM correctly identifies the target image. It is formally defined as:
% \begin{equation}
%     A = \alpha*\frac{1}{N} \sum_{n=1}^{N} \text{I}\left(i_n = \hat{i}_n\right),
% \end{equation}
% where $N$ denotes the total number of evaluation instances, $i_n$ is the ground-truth target image, and $\hat{i}_n$ is the prediction of the model. The indicator function $\text{I}(\cdot)$ equals 1 if the condition is satisfied and 0 otherwise. The correction factor $\alpha = \frac{1}{\text{Reasonable Rate}}$ represents human-verified adjustments based on human labels; the more errors in response generation, the bigger it is. In this way, the accuracy can be less influenced by the mistakes from the Guess-Agent. 
% The details can be found in Appendix~\ref{page:alpha}.

\paragraph{Response Accuracy} evaluates whether the MLLM correctly identifies the target image. It is formally defined as the average accuracy across $N$ evaluation sessions, adjusted by a correction factor $\alpha$ as:
\begin{equation}
\small
    \mathcal{A} = \alpha \cdot \frac{1}{N}\sum\limits_{n=1}^N \mathbf{I} (\hat{i}_n = i_n^{\prime}).
\end{equation}
The indicator function $\mathbf{I}(\cdot)$ equals \(1\) if the condition is satisfied and \(0\) otherwise. The correction factor $\alpha = 1/R_\text{agent}$ represents human-verified adjustments based on human labels. The term $R_\text{agent}$ (GuessAgent reliability) measures the rate at which the GuessAgent provides a correct and valid response. a lower $R_\text{agent}$ results in a larger $\alpha$, ensuring the accuracy score is less influenced by the external mistakes from the GuessAgent.

% \textbf{Reasoning efficiency} evaluates how quickly the MLLM completes the reasoning process. It is defined as:
% \begin{equation}
% R = \frac{1}{\text{exp}(\beta \frac{l - L_{\text{min}}}{L_{\text{max}} - L_{\text{min}}})},
% \end{equation}
% where $\beta$ is the reward coefficient with set to 1 in our experiments, and $l$ denotes the number of reasoning steps. The candidate set size is denoted as $B$. Following information-theoretic principles, we use $\log_2(B)$ to represent the minimal reasoning length $L_{\text{min}}$, and set the maximum length $L_{\text{max}}$ to 10. In this way, models that complete the reasoning process with fewer steps receive higher scores, thereby reflecting greater efficiency in active reasoning.

\paragraph{Reasoning Efficiency} quantifies how quickly the MLLM completes the reasoning process, rewarding models that reach the solution in fewer steps. It is modeled as:
\begin{equation}
\small
    \mathcal{R} = \frac{1}{N} \sum\limits_{n=1}^N \text{exp}(-\beta \frac{T_n - T_{\text{min}}}{T_{\text{max}} - T_{\text{min}}}),
\end{equation}
where $\beta$ is a reward coefficient, set to \(1\) in our experiments, and $T_n$ denotes the actual number of interaction length taken in the $n$-th evaluation session. The minimal reasoning length, $T_\text{min} = \lceil \log_2 B\rceil$, is derived from information-theoretic principles (ideal binary search), where $B$ is the candidate set size and $\lceil \cdot \rceil$ denotes the ceiling operator. We set the practical maximum length $T_\text{max}$ to 10. Consequently, models that complete the task with a query length $T$ closer to $T_\text{min}$ receive higher scores, directly reflecting greater efficiency in active reasoning progress.

% \textbf{Planning competence} measures a model’s ability to formulate effective strategies based on acquired information. It is defined as:
% \begin{equation}
% P = \text{exp}\left(-\max\left(0, \frac{l - L_B}{L_B}\right)\right),
% \end{equation}
% where $L_B$ denotes the maximum reasoning length under a one-by-one querying strategy, which we set equal to the size of the candidate set $B$. If the actual reasoning length $l$ exceeds $L_B$, the model is penalized, as this indicates a failure to re-plan effectively given the newly acquired evidence.

\paragraph{Planning Competence} measures the model's ability to formulate effective, adaptive strategies based on acquired evidence. It is defined using an exponential penalty function:
\begin{equation}
\small
    \mathcal{P} = \frac{1}{N} \sum\limits_{n=1}^N \exp\left(-\max\left(0, \frac{T_n - T_B}{T_B}\right)\right),
\end{equation}
where $T_B$ is the upper bound implied by a simple one-by-one querying strategy, instantiated as the candidate-pool size $B$ (i.e., $T_B=B$). If the actual reasoning step $T$ exceeds $T_B$, the term inside the $\max(\cdot)$ function becomes positive, resulting in an exponential penalty. This formulation specifically penalizes models whose active reasoning steps exceed the limit of a non-strategic, exhaustive search, indicating a failure to re-plan effectively using the newly acquired evidence.

%% file: page/04Experiments.tex
% \section{Experiments}

\section{Experimental Settings}
\paragraph{Data Distribution}
We conduct evaluations using 1500 real-world images and 500 synthetic images for each synthetic domain. For each experiment, we conduct $N=100$ evaluation sessions with a candidate pool size $B$ of 8.

\paragraph{Evaluation Models}
\input{table/model}

\input{table/main}

To comprehensively investigate the factors that influence active reasoning performance, we evaluate 20 mainstream MLLMs of varying sizes, architectures, and training strategies. Since the reasoning mode is closely tied to active reasoning performance, we report it separately. The details are summarized in Table~\ref{tab:model}.

\paragraph{Enhanced  Strategies}
\label{app-enhance}
The enhanced strategies are divided into two categories: perception-enhanced and thinking-enhanced approaches. The former includes RAG~\cite{chen2025svragloracontextualizingadaptationmllms} and the reasoning prompt~\cite{yu-etal-2023-exploring}, while the latter comprises CoT~\cite{wei2023chainofthoughtpromptingelicitsreasoning} and ReAct~\cite{Yao2022ReActSR}. Specifically, RAG provides retrieval results based on external information, and the reasoning prompt encourages the model to focus on specific aspects of the experimental setting. In contrast, CoT guides the model to reason step by step, whereas ReAct enables the model to plan subsequent actions as new information is obtained. The detailed prompts are shown in Appendix~\ref{app-prompt}.

% \paragraph{Enhanced  Strategies}
% \label{app-enhance}
% The enhanced strategies are divided into two categories: perception-enhanced and reasoning-enhanced approaches. The former includes RAG~\cite{chen2025svragloracontextualizingadaptationmllms} and the reasoning prompt~\cite{yu-etal-2023-exploring}, while the latter comprises CoT~\cite{wei2023chainofthoughtpromptingelicitsreasoning} and ReAct~\cite{liao2025reflectoolreflectionawaretoolaugmentedclinical}. Specifically, RAG provides retrieval results based on external information, and the reasoning prompt encourages the model to focus on specific aspects of the experimental setting. In contrast, CoT guides the model to reason step by step, whereas ReAct enables the model to plan subsequent actions as new information is obtained.

\section{Main Results}
The main results on different fields of images are presented in Table~\ref{tab:main}. Several key conclusions can be drawn from these results.

\paragraph{\textit{Existing MLLMs lack the ability to proactively obtain missing evidence.}}
Performance on active visual reasoning is weak across image fields. More than half of the models yield means below \(0.4\), indicating poor ability to retrieve target images within the required steps.
Even the strongest models, Qwen2.5-VL-72B and Qwen3-VL-Thinking, averages only \(0.70\) and \(0.65\).
% Even the strongest models, Qwen2.5-VL-72B and GPT-4o-mini, averages only \(0.7\) and \(0.5\).
The gap to reliable performance remains large, and further improvements are required.

% 这个
\paragraph{\textit{Synthetic images are more challenging.}}
Across categories, synthetic images yield lower scores than real-world images. The real-world images score the highest among various models, with \(0.7695\) for Qwen2.5-VL-32B and \(0.6844\) for Mimo-VL, while synthetic categories trail. GPT-4o-mini achieves \(0.5909\) on real-world images but only \(0.2972\) on kitchen. We hypothesize that synthetic domains typically exhibit repeated layouts and subtle semantic variations, creating higher demands in visual perception and world knowledge. These results indicate a need to enhance these two capabilities, particularly the recognition and utilization of fine-grained differences.

% \paragraph{\textit{Perception-based is worse than knowledge-based}}
\paragraph{\textit{Perception tasks underperform knowledge tasks.}}
Among all synthetic image categories, the performance on perception-based tasks is consistently lower than that on knowledge-oriented tasks. The average score for perception tasks is \(0.2966\), while for knowledge-oriented tasks it is \(0.4090\), indicating a relative difference of approximately \(27.5\%\). The most significant disparity is observed with Intern3-VL-14B in the traffic and kitchen scenarios, where the scores are \(0.6395\) and \(0.1915\) respectively, resulting in a substantial performance gap.
Our hypothesis is that perception tasks focus on fine-grained visual recognition, whereas knowledge-oriented tasks emphasize knowledge inference and assumption. These results indicate that the models exhibit stronger capabilities in knowledge reasoning compared to visual perception.
% Perception tasks focus on fine-grained visual recognition, whereas knowledge-oriented tasks emphasize knowledge inference and assumption. These results indicate that the models exhibit stronger capabilities in knowledge reasoning compared to visual perception, and need to improve more on it.

\paragraph{\textit{Scaling generally improves performance, but not monotonically.}}
A key trend is the overall benefit of scaling: larger parameter sizes deliver substantial gains across model families. In the Qwen2.5-VL series, the average score rises from \(0.1523\) (3B) to \(0.6978\) (72B), representing more than \(300\%\) improvement, also happened in Intern3-VL-series.
% Inter3-VL follows a similar trajectory, increasing from \(0.0985\) (2B) to \(0.4125\) (38B). 
These results reinforce scaling as a reliable means of enhancing both perceiving and thinking~\cite{Chen2023PaLIXOS}, echoing two processes in active reasoning. However, the trend is not strictly monotonic: Qwen2.5-VL-32B achieves an average score of \(0.5894\), clearly outperforming Intern3-VL-38B at \(0.4125\) despite their similar scale, suggesting that beyond a certain scale, the stage and quality of training constrain both perception and thinking.

% \paragraph{\textit{Explicit thinking can effectively improve it.}}
\paragraph{\textit{Explicit thinking significantly boosts performance.}}
Across various models, the integration of explicit thinking mechanisms consistently leads to substantial performance improvements. For example, Kimi-VL demonstrates this clearly: its base model achieves an average score of only \(0.2201\), while the thinking-enhanced variant, Kimi-VL-Thinking, reaches \(0.3153\), an increase of \(43.3\%\). Notably, smaller models augmented with thinking capabilities can rival or even surpass much larger models without such mechanisms. Mimo-VL (\(0.6102\)), trained with explicit thinking reinforcement learning from Qwen2.5-VL-7B, outperforms GPT-4o-mini (\(0.5050\)). These findings suggest that thinking augmentation is an effective pathway to stronger active reasoning, especially in dynamic information settings.

%% file: table/model.tex
\begin{table}[t]
  \center
  \begin{adjustbox}{max width=\columnwidth, center}
    \begin{tabular}{lccc}
      \toprule
      \textbf{Model Name} & \textbf{Thinking Mode} & \textbf{Parameters} \\ 
      \midrule
      Intern3-VL~\citeyearpar{zhu2025internvl3} & Non-think & 2B, 8B, 14B, 38B, 78B  \\
      Intern3.5-VL~\citeyearpar{wang2025internvl3_5} & Both & 2B, 4B, 8B, 14B, 38B  \\
      Qwen-2.5-VL~\citeyearpar{bai2025qwen25vltechnicalreport} & Non-think & 3B, 7B, 32B, 72B \\
      Qwen-3-VL-Instrust~\citeyearpar{qwen3-vl-blog} & Non-think & 30B-A3B \\
      Qwen-3-VL-Thinking~\citeyearpar{qwen3-vl-blog} & think & 30B-A3B \\
      Mimo-VL~\citeyearpar{coreteam2025mimovltechnicalreport} & Think & 7B  \\
      Kimi-VL~\citeyearpar{kimiteam2025kimivltechnicalreport} & Non-Think & 16.4B-A3B   \\
      Kimi-VL-Thinking~\citeyearpar{kimiteam2025kimivltechnicalreport}  & Think & 16.4B-A3B  \\
      GPT4o-mini~\citeyearpar{openai2024gpt4omini} & Non-think & - \\
      \bottomrule
    \end{tabular}
  \end{adjustbox}
    \caption{Details of evaluation MLLMs.}
  \label{tab:model}
  \vspace{-1.5em}
\end{table}

%% file: table/main.tex
\begin{table*}[htbp]
\centering
\begin{adjustbox}{max width=0.95\textwidth, center}
\begin{tabular}{l *{13}{S[table-format=1.4]}}
\toprule
\multicolumn{1}{c}{\multirow{2}{*}{\textbf{Model}}} &
\multicolumn{1}{c}{\multirow{2}{*}{\textbf{Real}}} &
\multicolumn{5}{c}{\textbf{Perception}} &
\multicolumn{4}{c}{\textbf{Knowledge}} &
\multicolumn{3}{c}{{\textbf{Average}}} \\
\cmidrule(lr){3-7} \cmidrule(lr){8-11} \cmidrule(lr){12-14}
& & \textbf{City} & \textbf{Face} & \textbf{Kitchen} & \textbf{Livingroom} & \textbf{Nature} &
\textbf{Animal} & \textbf{Traffic} & \textbf{Vehicle} & \textbf{Occupation} & \textbf{Per} & \textbf{Know} & \textbf{Total}\\
\midrule
\multicolumn{14}{c}{\textbf{Inter3-VL-Series}} \\
\midrule
Intern3-VL-2B & 0.0896 & 0.1009 & 0.1430 & 0.1224 & 0.1196 & 0.1118 & 0.0941 & 0.0390 & 0.0484 & 0.1162 &0.1195&0.0744& 0.0985 \\
Intern3-VL-8B & 0.2626 & 0.1547 & 0.2621 & 0.1759 & 0.0851 & 0.2237 & 0.3244 & 0.4004 & 0.2162 & 0.3263 & 0.1803&0.3168&0.2423 \\
Intern3-VL-14B & 0.4247 & 0.2522 & 0.2364 & 0.1915 & 0.2690 & 0.2643 & 0.4392 & 0.6395 & 0.4658 & 0.5082 & 0.2427&0.5131&0.3691 \\
Intern3-VL-38B & 0.5294 & 0.3000 & 0.3864 & 0.3057 & 0.2364 & 0.3787 & 0.4642 & 0.5191 & 0.4690 & 0.5360 & 0.3214&0.4970&0.4125 \\
Intern3-VL-78B & 0.4332 & 0.2056 & 0.2154 & 0.2923 & 0.2455 & 0.3086 & 0.4708 & 0.3884 & 0.3705 & 0.3825 &0.2535&0.4031& 0.3313 \\
\midrule
\multicolumn{14}{c}{\textbf{Qwen2.5-VL-Series}} \\
\midrule
Qwen2.5-VL-3B & 0.1496 & 0.0958 & 0.1459 & 0.0865 & 0.0888 & 0.1151 & 0.2132 & 0.1924 & 0.2140 & 0.2214 &0.1064&0.2103& 0.1523 \\
Qwen2.5-VL-7B & 0.5087 & 0.1761 & 0.1509 & 0.1844 & 0.1910 & 0.2544 & 0.4863 & 0.4718 & 0.5233 & 0.5196 &0.1914&0.5003& 0.3466 \\
Qwen2.5-VL-32B & \underline{0.7695} & 0.5284 & 0.5980 & 0.3717 & 0.5205 & 0.5274 & 0.6727 & 0.5237 & \underline{0.6735} & 0.7087 &0.5092&0.6446& 0.5894 \\
Qwen2.5-VL-72B & \textbf{0.7722} & \underline{0.5655} & \textbf{0.6594 }& \textbf{0.5700} & \textbf{0.6570} & \underline{0.6445} & \textbf{0.7684} & \textbf{0.8177} & \textbf{0.7263} & \textbf{0.7968} &\textbf{0.6193}&\textbf{0.7773}& \textbf{0.6978} \\
\midrule 
\multicolumn{14}{c}{\textbf{Inter3.5-VL-Series}} \\
\midrule
Intern3.5-VL-2B & 0.1226 & 0.1316 & 0.0994 & 0.0999 & 0.0998 & 0.0611 & 0.1136 & 0.0496 & 0.1053 & 0.1030 & 0.0983&0.0929&0.0986 \\
Intern3.5-VL-4B & 0.3977 & 0.1560 & 0.2015 & 0.1782 & 0.1462 & 0.1299 & 0.3158 & 0.2018 & 0.2274 & 0.1774 & 0.1623&0.2306&0.2132 \\
Intern3.5-VL-8B & 0.4048 & 0.1096 & 0.3464 & 0.2253 & 0.2093 & 0.1144 & 0.2071 & 0.2560 & 0.2047 & 0.2125 & 0.2010&0.2201&0.2290 \\
Intern3.5-VL-14B & 0.1675 & 0.2082 & 0.2085 & 0.2193 & 0.1501 & 0.2397 & 0.2193 & 0.1704 & 0.1683 & 0.2505 & 0.2052&0.2021&0.2002 \\
Intern3.5-VL-38B & 0.2304 & 0.2150 & 0.3203 & 0.1994 & 0.1320 & 0.1860 & 0.2878 & 0.3307 & 0.3058 & 0.3280 & 0.2105&0.3131&0.2535 \\
\midrule
\multicolumn{14}{c}{\textbf{Others}} \\
\midrule
Kimi-VL & 0.2957 & 0.1296 & 0.2259 & 0.1570 & 0.1773 & 0.1773 & 0.3689 & 0.2115 & 0.2799 & 0.1778 &0.1734&0.2595& 0.2201 \\
Kimi-VL-Thk. & 0.3687 & 0.2567 & 0.2807 & 0.2703 & 0.2860 & 0.3569 & 0.3366 & 0.3532 & 0.3309 & 0.3131 &0.2901&0.3334& 0.3153 \\
\hdashline
Qwen3-VL-Ins & 0.6935 & 0.4524 & 0.5594 & 0.3849 & 0.4825 & 0.5425 & 0.6239 & 0.6753 & 0.6106 & 0.5605 &0.4843   &0.6176   & 0.5586 \\
Qwen3-VL-Thk & 0.7017  & \textbf{0.6664}  &  0.6105 & 0.4478  &\underline{0.6395}   & \textbf{0.7176}  & 0.6779  & \underline{0.7071}  & 0.5932  & \underline{0.7724}  &  \underline{0.6164} & \underline{0.6877}  & \underline{0.6534}  \\ 
\hdashline
Mimo-VL & 0.6844 & 0.4492 & \underline{0.6202} & \underline{0.4811} & 0.5607 & 0.6043 & \underline{0.7221} & 0.6726 & 0.6017 & 0.7061& 0.5431 &0.6756 & 0.6102 \\
GPT-4o-mini   & 0.5909 & 0.4235 & 0.4473 & 0.2972 & 0.4105 & 0.4410 & 0.6342 & 0.6070 & 0.6068 & 0.5918 & 0.4039 & 0.6100 & 0.5050 \\
\midrule
\multicolumn{14}{c}{\textbf{Average}} \\
\midrule
 \multicolumn{1}{c}{--} & 0.4299  & 0.2789  & 0.3359  & 0.2631  &0.2853   & 0.3199  &0.4220   &0.4114   &0.3871   &0.4154   &  0.2966 & 0.4090  & 0.3549  \\
\bottomrule
\end{tabular}
\end{adjustbox}
\caption{Model performance across different categories. The Per and Know are abbreviations for perception and knowledge domains. Ins and Thk are abbreviations for instruct and thinking. The best scores are in \textbf{bold}, and the secondary are \underline{underlined}.}
\label{tab:main}
\vspace{-1.5em}
\end{table*}

%% file: page/05Analysis.tex
\section{Further Discussions}
% We make further exploration on the deep factors of multimodal active reasoning, and the results are shown in Figure~\ref{fig:edit}-Figure~\ref{fig:ana-thinking}. 
We delve deeper into the underlying factors of multimodal active reasoning, with the analysis results presented in {Figure~\ref{fig:passive_active} to Figure~\ref{fig:ana-thinking}.

\subsection{Active Reasoning is Effective but Limited}
\begin{figure}[t]
    \centering
    \includegraphics[width=0.85\linewidth]{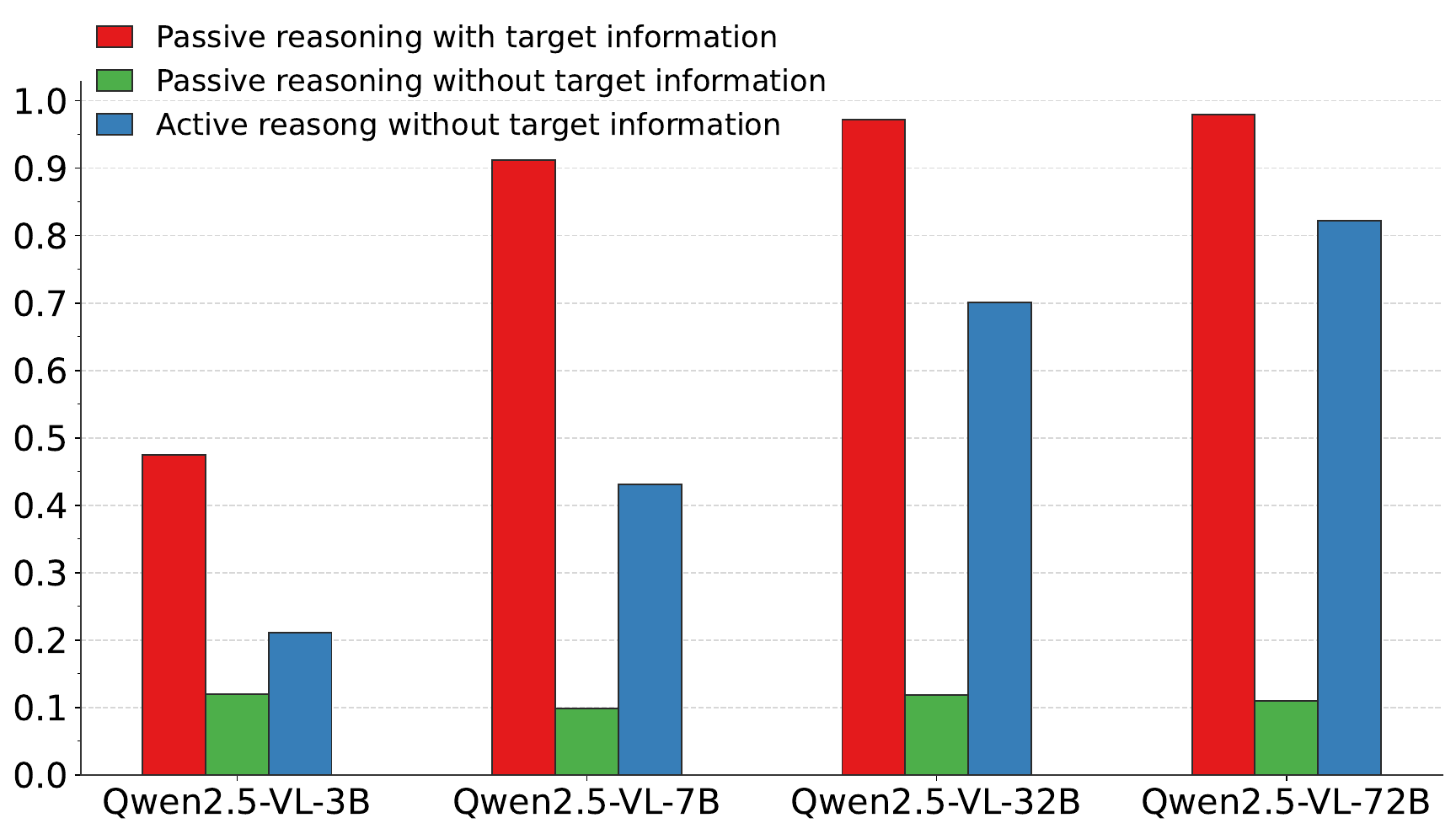}
    \caption{The accuracy compared with the passive reasoning. The no-information baseline yields near-chance accuracy, confirming the reliability of our tasks.}
    \label{fig:passive_active}
    \vspace{-1.5em}
\end{figure}

% \paragraph{\textit{Active reasoning can recover missing information.}}
% To compare active and passive reasoning, we run experiments under identical settings and provide a target image description only in the passive condition, thereby shifting evidence from partial to complete.  Figure~\ref{fig:passive_active} shows that passive reasoning with a target description attains accuracy near \(1\), with larger models benefiting from stronger perception and information investigation. Removing the target description causes a drop of more than \(0.72\) points on average, underscoring the decisive role of task-specific information. Under the active setting, although no target description is provided, accuracy rises to about \(0.54\) on average and improves with model size, approaching the fully informed passive condition. This pattern indicates that active reasoning can acquire the missing evidence and therefore enables correct decisions. 

\paragraph{\textit{Active reasoning may exhibit two failure modes.}}
\label{subsec:active}
To validate the efficacy of active reasoning, we established a passive reasoning baseline under identical settings by providing the complete target image description, thereby moving the evidence from partial to complete. Figure~\ref{fig:passive_active} demonstrates that performance under passive reasoning is strongly size-dependent: large models achieve high accuracy ($>0.9$), while the smaller 3B model yields poor accuracy ($<0.5$) even with the full target information. This size-dependent performance on passive reasoning suggests two potential failure modes for active reasoning: small models struggle with integrating multimodal information, while larger models that perform well here may still fail in the active setting by failing to ask the right questions for information seeking incompletely.
% and integrate historical information from the dialogue.

\paragraph{\textit{Active reasoning can recover missing information.}}
As Figure~\ref{fig:passive_active} shows, removing the target description causes a drop of more than \(0.72\) points on average, underscoring the decisive role of task-specific information. Under the active setting, although no target description, accuracy rises to about \(0.54\) on average and improves with model size, approaching the fully informed passive condition. This pattern indicates that active reasoning can acquire the missing evidence and therefore enables correct decisions.
% 0.5415， 0.8345  0.1115

\paragraph{\textit{The ability of active reasoning is far from passive reasoning.}}
Figure~\ref{fig:passive_active} further shows that the gap between active reasoning and passive reasoning with target information exceeds idealized expectations, below \(0.29\) on average. The gap becomes smaller as model size increases, yet active performance does not reach the level of passive reasoning. This persistent shortfall is attributable to limited information seeking and integration during interaction, resulting in not acquiring enough information. Hence, there remains considerable room for improvement in active reasoning.

\subsection{Perception is Crucial for Active Reasoning}
\paragraph{\textit{Fine-grained difference is more difficult for models to perceive.}}
\begin{figure}[t]
    \centering
    \includegraphics[width=0.85\linewidth]{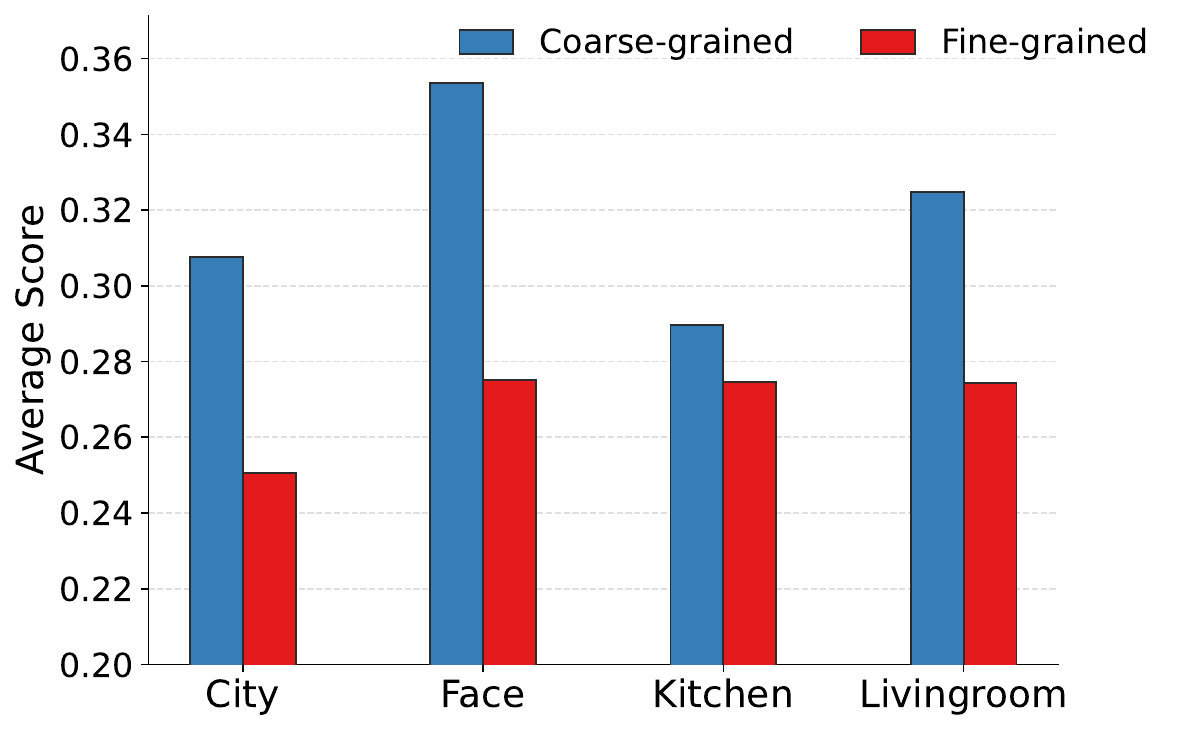}
    \caption{The difference between the original images (coarse-grained) and edited images (fine-grained).}
    \label{fig:edit}
    \vspace{-1.5em}
\end{figure}
To further examine the performance of active reasoning with the perception capability, we construct new image sets across four perception-oriented domains by editing a single attribute of the original images, as shown in Appendix~\ref{app-sec:edit_case}. As shown in Figure~\ref{fig:edit}, the scores of almost all models decrease substantially relative to those on the original sets, with the degradation of \(15.8\%\) on average. This pattern indicates that variation in the candidate pools strongly affects model performance, underscoring the importance of perceiving the candidate pool at the beginning.
%  of the task.
% 类型	平均得分
% Normal	0.3189
% Edit	0.2686

\begin{figure*}[t]
    \centering
    \begin{minipage}[c]{0.48\textwidth}
        \centering
        \includegraphics[width=0.85\textwidth]{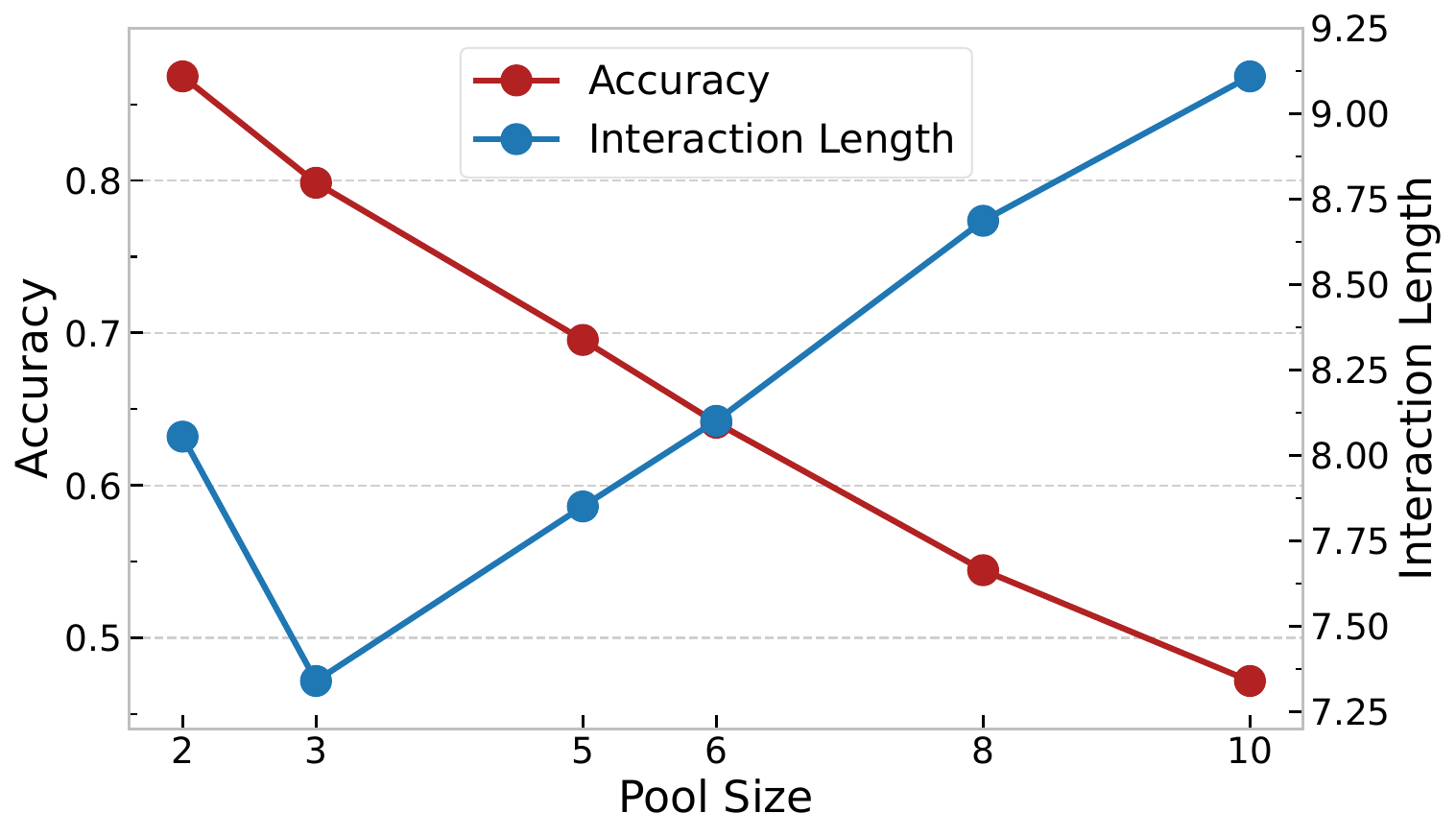}
        \subcaption{Accuracy and interaction length with different pool sizes.}
        \label{fig:acc_lens_bs}
    \end{minipage}
    \hspace{0.01\textwidth}
    \begin{minipage}[c]{0.48\textwidth}
        \centering
        \includegraphics[width=0.8\textwidth]{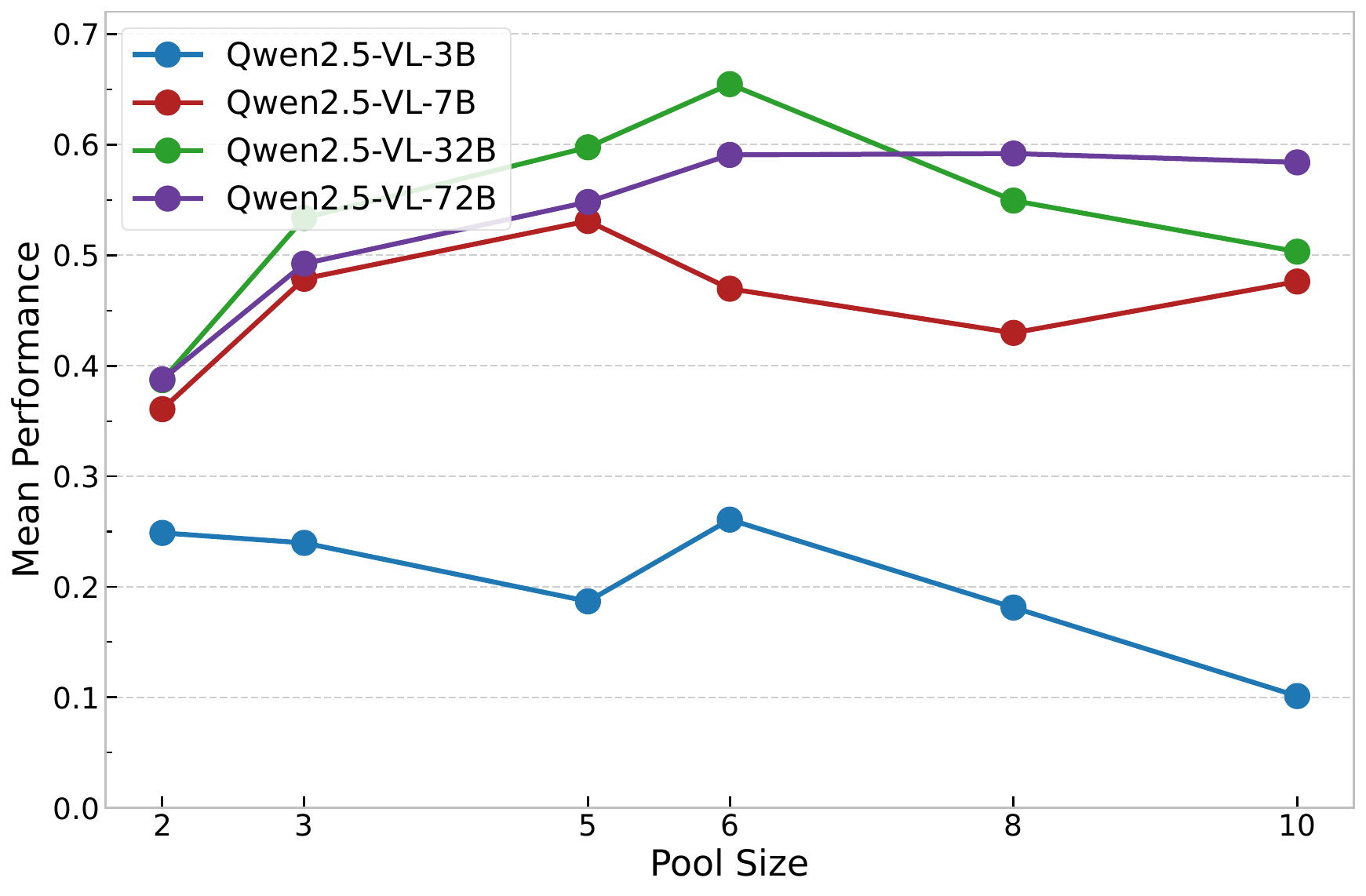}
        \subcaption{Model with different pool sizes.}
        \label{fig:score-diff-size}
    \end{minipage}

    \caption{The performance under different models and pool sizes. The left part is the accuracy and interaction length of four models on average, and the right part is the average scores of four models.}
    \label{fig:bs_image}
    \vspace{-1.5em}
\end{figure*}

\paragraph{\textit{More images lead to worse accuracy and longer steps.}}
To examine how perceptual capability affects overall performance as a function of candidate pool size, we evaluate Qwen2.5-VL-series models under varying numbers of candidate images, as illustrated in Figure~\ref{fig:acc_lens_bs}. As pool size increases, accuracy generally declines while the number of reasoning steps rises, with a relative \(45.7\%\) degradation in accuracy and a \(13.1\%\) increase in steps on average. Larger pools place higher demands on perception, which in turn strongly influences the thinking process. These results suggest that perception is central to active reasoning: models that can identify and focus on a small, relevant subset of the image pool can sustain effective reasoning.
% 2	0.8682	8.0550
% 3	0.7985	7.3402
% 5	0.6955	7.8508
% 6	0.6412	8.1012
% 8	0.5445	8.6863
% 10    0.4718	9.1086

% \paragraph{Model always gets the best performance on a medium number.}
\paragraph{\textit{Unbalanced between perception and thinking limits the performance.}}
Figure~\ref{fig:score-diff-size} reports overall scores across models and shows that all models reach their peak performance at a moderate pool size. At larger pool sizes, performance declines by about \(15.8\%\) on average, driven by substantial drops in accuracy that are attributable to perceptual limitations. With small pools, results fall to roughly \(70.1\%\) of the moderate-pool performance, primarily due to limited thinking capacity, which is penalized by planning competence for redundant interactions. These observations support the reliability of our metric that considers both accuracy and efficiency. Moreover, the results indicate that a good balance between perception and thinking is essential for strong performance.

% Batch_Size (BS)	平均得分
% 2	0.3460
% 3	0.4360
% 5	0.4658
% 6	0.4939
% 8	0.4380
% 10	0.4161

% In this setting, the degree of missing evidence increases with the number of images: as more images are introduced, each individual query contains only a partial view of the overall information, thereby amplifying the challenge of evidence completion. 

\subsection{Unreliable Verification Limits Gains}
\paragraph{\textit{Unreliable verification lengthens the reasoning process}}
\begin{figure}
\centering
\includegraphics[width=0.85\linewidth]{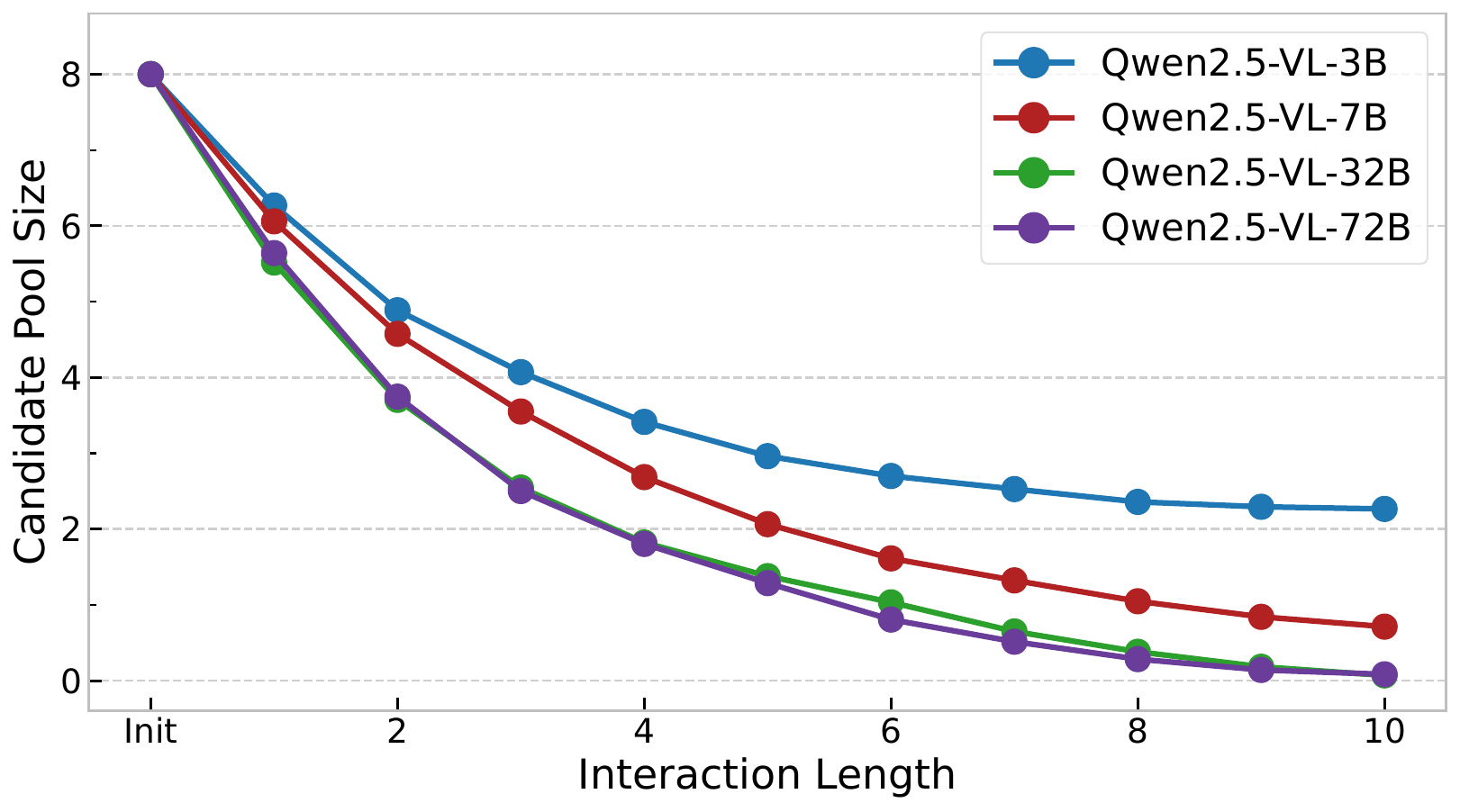}
\caption{Candidate image–pool size versus stepwise reasoning progress. We use Qwen2.5-VL-7B as a QA-based filter to prune the pool via QA-pairs.}
\label{fig:infor-gains}
\vspace{-1em}
\end{figure}

As shown in Figure~\ref{fig:infor-gains}, the effectiveness of active reasoning exhibits diminishing returns across steps: early steps yield sharper gains than later ones. Across four models, the first four steps account for about \(77.3\%\) of the total progress, whereas the remaining six steps contribute only about \(22.7\%\). As the reasoning sequence grows longer, the marginal information gain decreases, indicating a limited ability to integrate evidence dynamically and acquire new information. These patterns suggest that unreliable verification triggers additional steps with little payoff, thereby lengthening the overall reasoning process.

% Model	起始值 (mean[0])	第五个值 (mean[4])	最终值 (mean[10])	差值占比
% Qwen2.5-VL-3B	8	3.4137	2.2657	79.98%
% Qwen2.5-VL-7B	8	2.6858	0.7135	72.93%
% Qwen2.5-VL-32B	8	1.8227	0.0705	77.90%
% Qwen2.5-VL-72B	8	1.8043	0.0845	78.27%

\paragraph{\textit{Benefits of early stopping with single-candidate pools.}}
\begin{figure}[t]
    \centering    \includegraphics[width=0.85\linewidth]{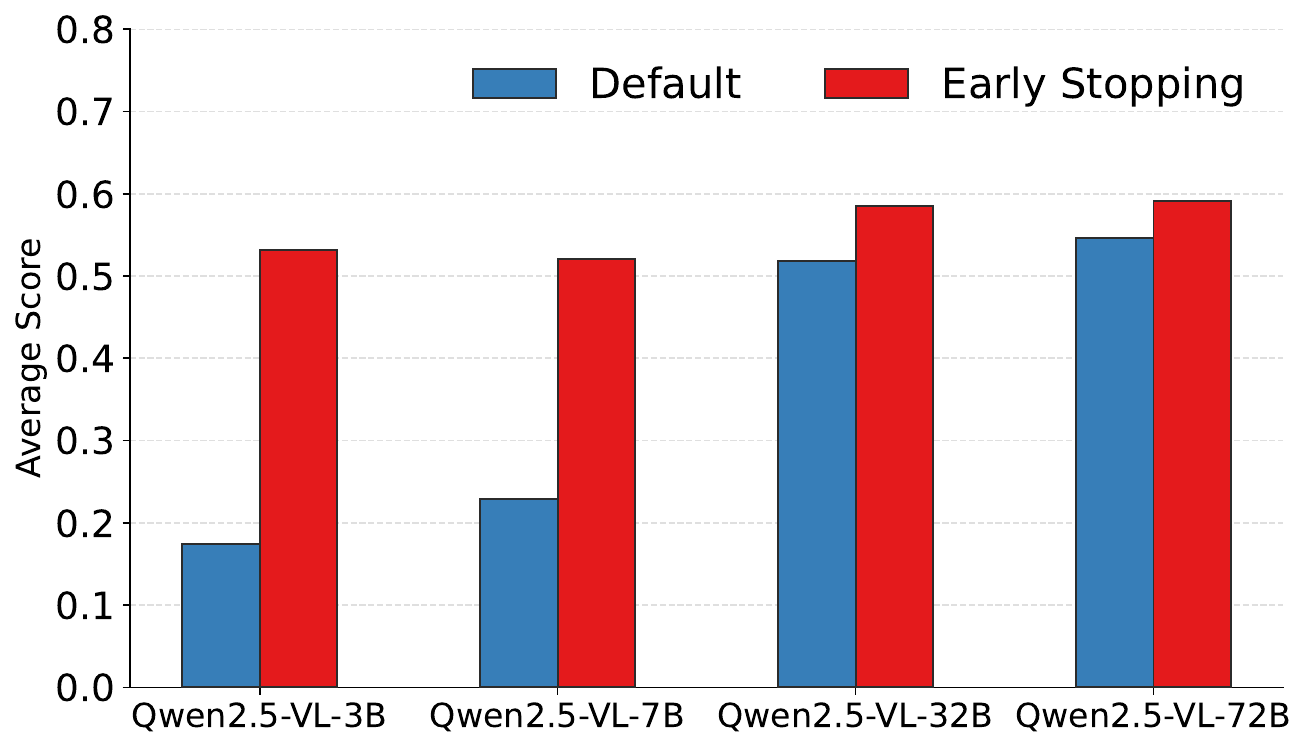}
    \caption{The different performance between the default process and the early stopping.}
    \label{fig:pool_size_1}
\vspace{-1.5em}
\end{figure}

Further shown in Figure~\ref{fig:infor-gains}, when the candidate pool size is \(1\), models often fail to commit and instead continue multi-turn queries with negligible information gain. To test whether early truncation helps, we recompute scores by stopping as soon as the pool shrinks to \(1\) and outputting the remaining candidate as the final answer. As shown in Figure~\ref{fig:pool_size_1}, all four models improve under early truncation, with an average improvement of \(51.7\%\). The gap becomes smaller for larger MLLMs, indicating that larger models are more able to decide in time than smaller ones. These findings suggest that adopting reliable verification and early stopping policies can reduce unnecessary steps and improve both accuracy and efficiency.
% 数据列表	均值 (Mean)
% original	0.3670
% processed	0.5571

% \paragraph{More questions yield diminishing returns.}
% \begin{figure}[t]
%     \centering
%     \includegraphics[width=0.85\linewidth]{figure/step_acc.pdf}
%     \caption{The accuracy and cumulative accuracy on different reasoning steps.}
%     \label{fig:step_acc}
% \end{figure}
% We test whether additional questioning improves performance by computing per-step and cumulative accuracy across reasoning steps. As shown in Figure~\ref{fig:step_acc}, the middle portion of the interaction attains the highest accuracy and contributes the largest marginal gain, over \(10\%\) of the total, while later steps add little. This indicates that models struggle to surface the key evidence even with ample interactions.

\subsection{Various Strategies for Improvement}
\begin{figure}[t]
    \centering
    \begin{subfigure}[t]{0.5\textwidth}
        \centering
        \includegraphics[width=0.85\linewidth]{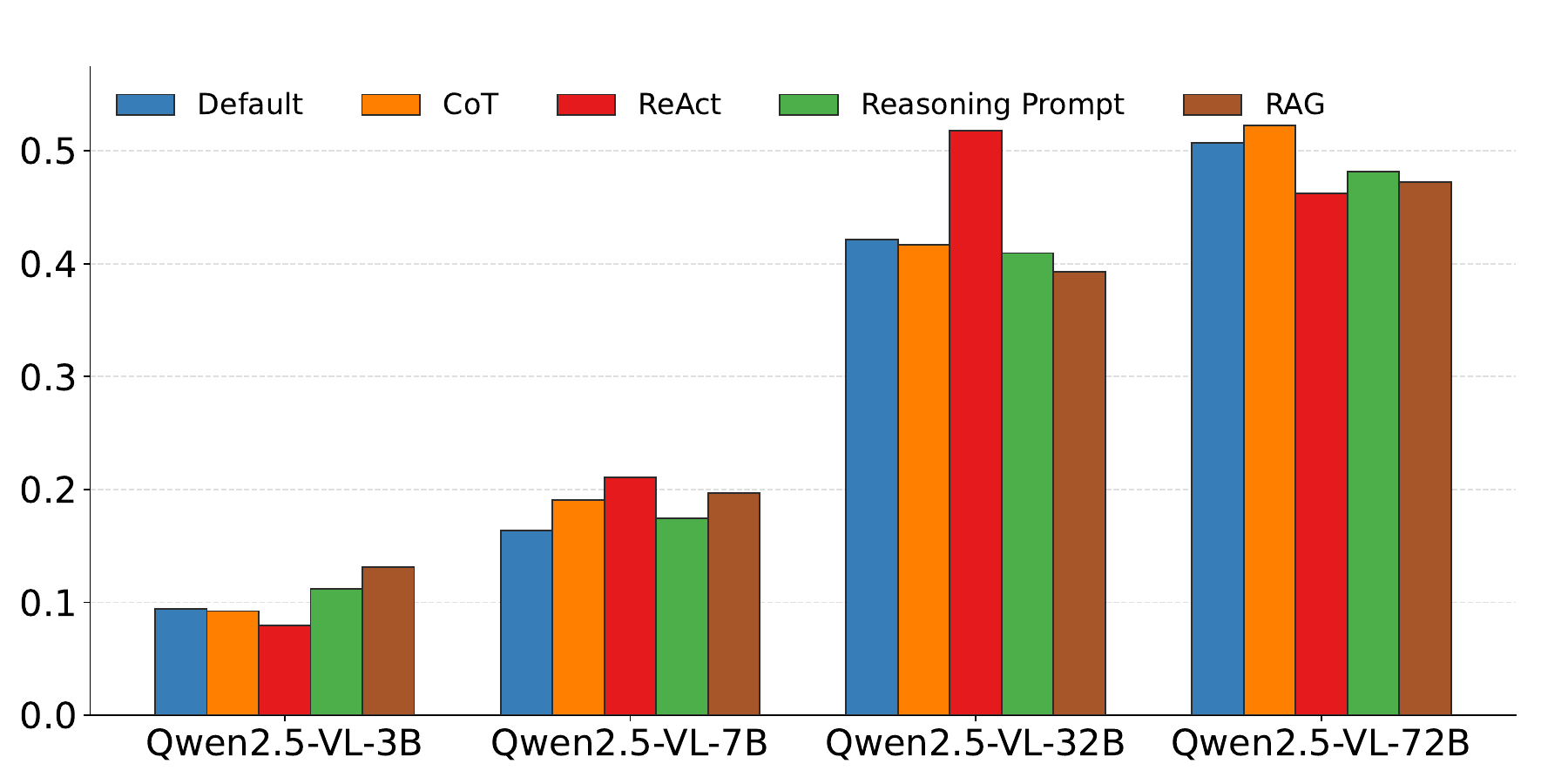}
        \caption{Performance in perception-oriented images.}
        \label{fig:ana-method-perc}
    \end{subfigure}

    % \vspace{0.1em}

    \begin{subfigure}[t]{0.5\textwidth}
        \centering
        \includegraphics[width=0.85\linewidth]{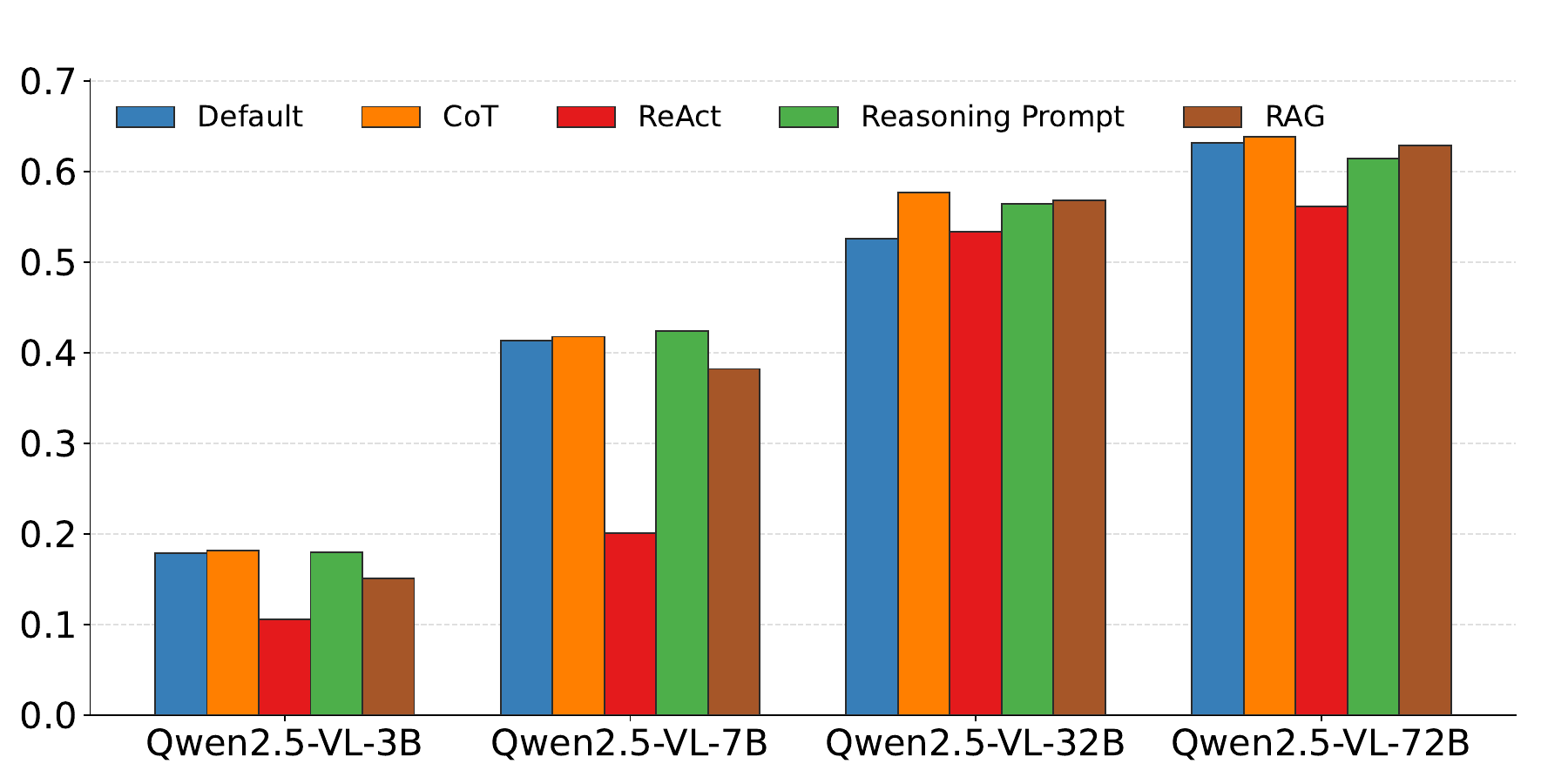}
        \caption{Performance in knowledge-oriented images.}
        \label{fig:ana-method-know}
    \end{subfigure}

    \caption{Performance under different strategies.}
    \label{fig:ana-method}
    \vspace{-1.5em}
\end{figure}

\paragraph{\textit{Perceptual enhancement yields larger gains than knowledge enhancement.}}
To further examine the effectiveness of different methods from the perspectives of perception and knowledge, we evaluate four enhancement strategies: two perception-enhanced methods (Reasoning Prompt and RAG) and two thinking-enhanced methods (CoT and ReAct), which are described in Section~\ref{app-enhance}. As shown in Figure~\ref{fig:ana-method}, the effects differ across task types. Interventions that strengthen perception improve performance on perception-oriented images, whereas knowledge-oriented interventions provide little additional benefit. These results suggest that current MLLMs already meet most knowledge requirements for active reasoning but still require stronger perceptual encoding and selection.

\paragraph{\textit{Different models require different improvements}}
As shown in Figure~\ref{fig:ana-method-perc}, the four methods have heterogeneous effects across models. Qwen2.5-VL-3B benefits most from perception-enhanced strategies, whereas the three larger models gain more from thinking-enhanced methods. This pattern suggests that smaller models are primarily constrained by perceptual capacity, while larger models profit more from strengthened reasoning, which proves the hypothesis in Section~\ref{subsec:active}. Improvements, therefore, should be model-specific, with training targeted to various dominant bottlenecks.

\paragraph{\textit{Training thinking modes are better than prompt-based thinking.}}
\begin{figure}[t]
    \centering
    \includegraphics[width=0.85\linewidth]{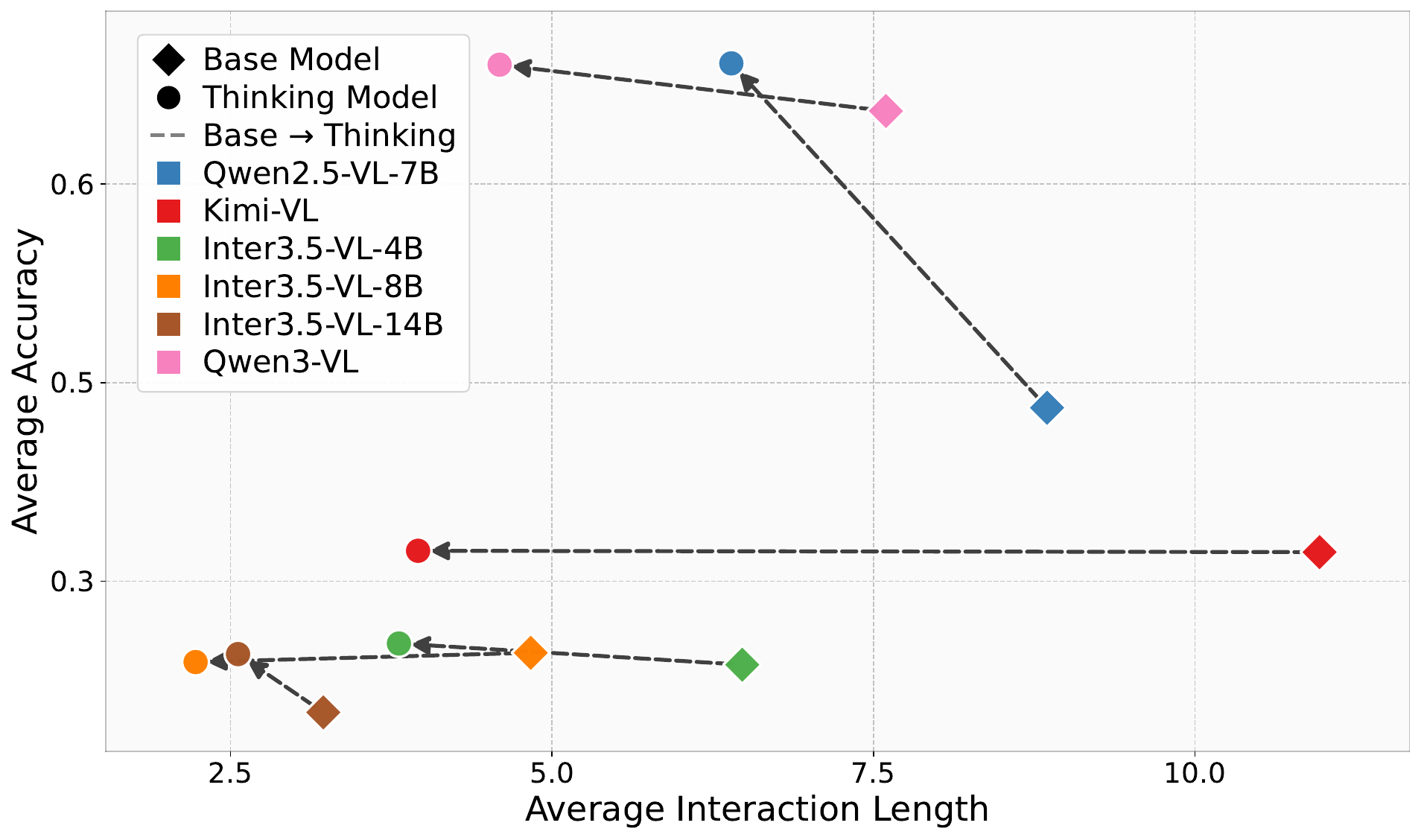}
    \caption{The accuracy and interaction length under thinking and non-thinking modes. The thinking mode of the Intern3.5-VL series is under the thinking prompt in official settings. }
    \label{fig:ana-thinking}
    \vspace{-1em}
\end{figure}
To further verify the effectiveness of explicit thinking, we evaluate multiple models in both non-thinking and thinking modes, with results shown in Figure~\ref{fig:ana-thinking}. Across all models, the thinking mode identifies more target images while using fewer reasoning steps than the nonthinking mode. The gains from the thinking mode also exceed those from training-free prompts such as CoT and ReAct. Notably, Mimo-VL, trained with thinking reinforcement learning from Qwen2.5-VL-7B, achieves a \(76.1\%\) improvement, substantially outperforming CoT and ReAct. These findings suggest that the training of explicit thinking enhances search and verification, motivating step-aware and verification-aware objectives.
% mimo 0.6102  qwen-7b 0.3466 CoT 0.3094

% These models consistently summarize available information and guide the direction of subsequent queries, thereby achieving significant performance gains. Moreover, the improvements brought by the thinking mode are larger than those obtained from CoT or ReAct enhancement strategies, indicating that training-free strategies offer limited improvements compared to the more substantial benefits achieved when models adopt the thinking mode after fine-tuning with SFT or RL. These findings provide a clearer understanding of how the thinking mode contributes to active reasoning, highlighting its stronger and more consistent benefits compared to training-free enhancement strategies.

%% file: page/06Conclusion.tex
\section{Conclusions}
To advance evaluation of multimodal active reasoning, we present \textsc{GuessBench}, a benchmark that assesses MLLMs on an interactive target guessing task. The setting requires active evidence acquisition and dynamic information integration, including both perception-oriented and knowledge-oriented images. Evaluating 20 leading MLLMs, we find that current systems cannot perform well, leaving substantial headroom for improvement. Our analysis identifies limited perceptual capability and delayed decisions as the main bottlenecks. Further experiments show that strengthening reasoning yields consistent gains. These findings point toward MLLMs that are better aligned with active multimodal tasks in real-world settings.

%% file: page/07Appendix.tex
\input{page/02Related_works}
\section{Verifying the reliability of the process.}
\input{table/human_label}
To verify whether the GuessAgent is able to provide appropriate information through question answering, we conduct a human verification of the entire evaluation procedure. Specifically, we randomly sample 100 cases from each model and annotate the responses with a binary label (True or False) to indicate whether the output is reasonable. The aggregated results are reported in Table~\ref{tab:human_rate}. As shown, all models achieve reasonable rates above 95\%, which indicates that our framework can accurately provide evidence in response to model queries,  thereby offering a reliable basis for evaluating models’ active reasoning ability.

\section{The details of the evaluation procedure.}
\label{app-prompt}
We provide the detailed prompts on the whole evaluation procedure and enhanced methods for reproduction, which are shown in Table~\ref{app-tab:prompt_mllm}-Table~\ref{app-tab:prompt_react}.
\input{table/prompt/base_prompt}
\input{table/prompt/guess_agent}
\input{table/prompt/method_prompt}

\section{Case Study}
\subsection{Dataset Cases}
\label{app-sec:edit_case}
\begin{figure*}[t]
    \centering
    \includegraphics[width=0.95\linewidth]{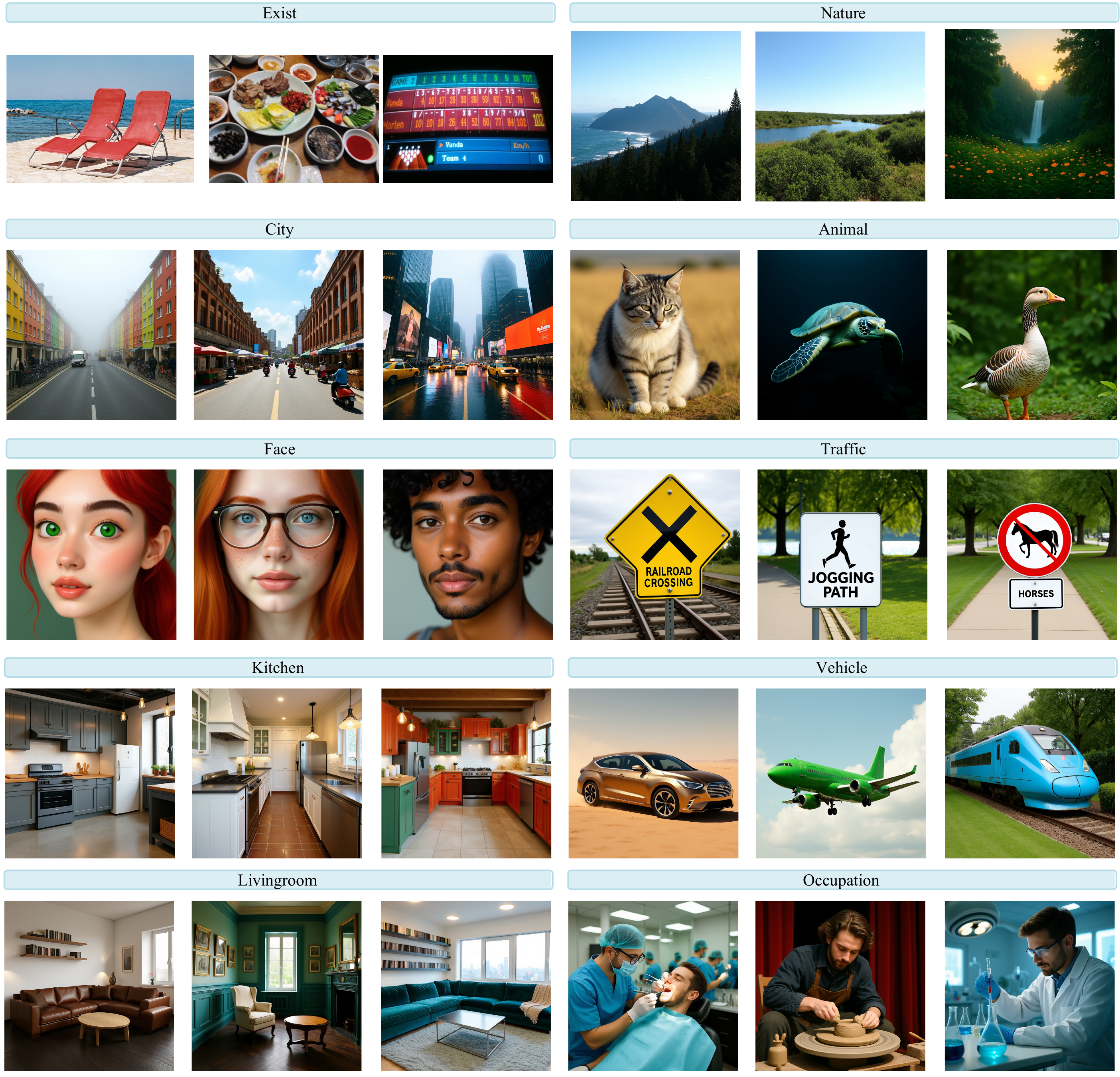}
    \caption{The cases of ten image fields.}
    \label{app-fig:dataset_case}
\end{figure*}

\begin{figure*}[t]
    \centering
    \includegraphics[width=0.95\linewidth]{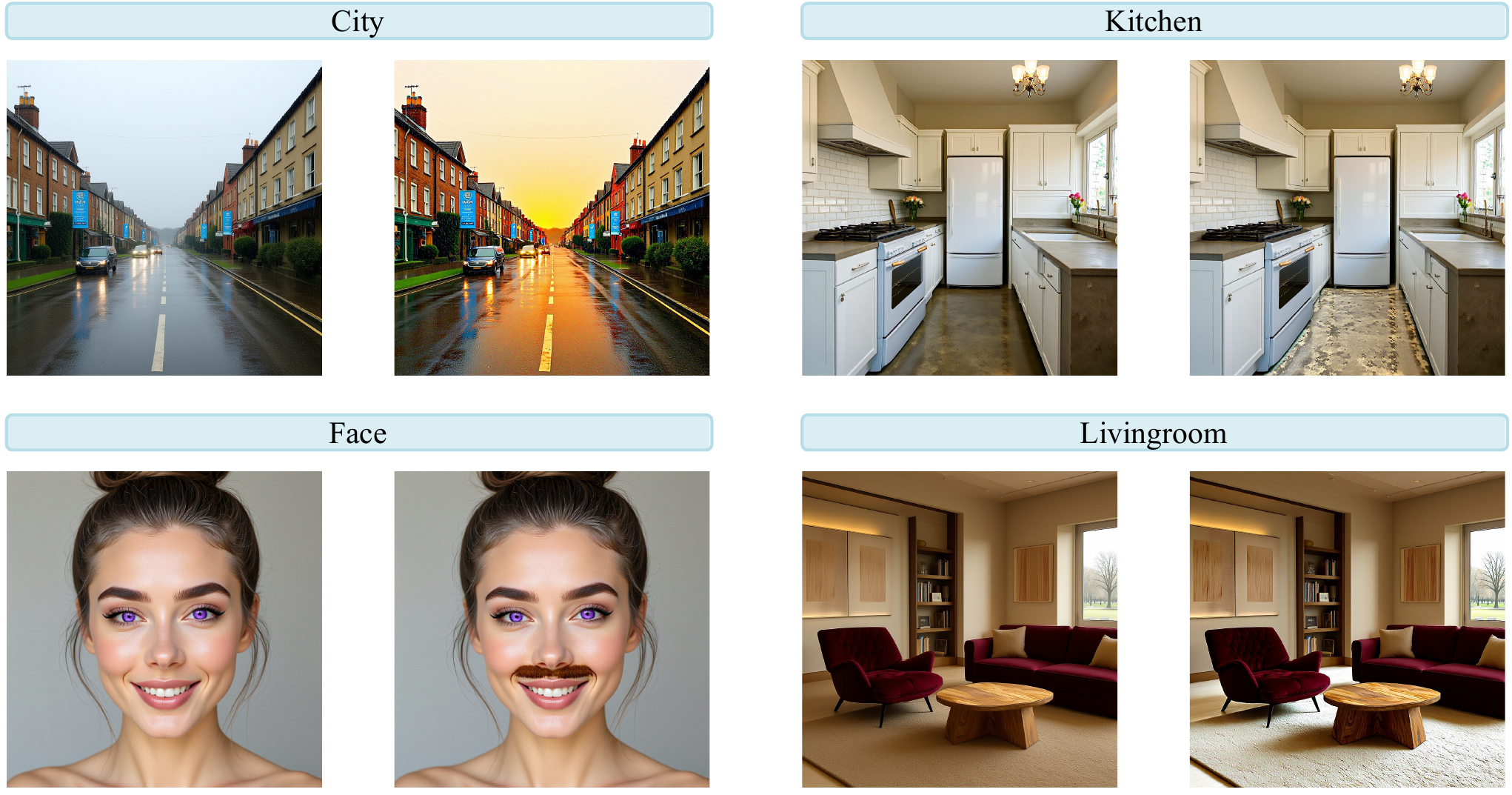}
    \caption{The cases of original images and edited images.}
    \label{app-fig:edit_case}
\end{figure*}

We provide examples of images in Figure~\ref {app-fig:dataset_case} and Figure~\ref {app-fig:edit_case}, including images in the main procedure and edited images in further experiments.

\subsection{Active Reasoning Cases}
\begin{figure*}[t]
    \centering
    \includegraphics[width=0.98\linewidth]{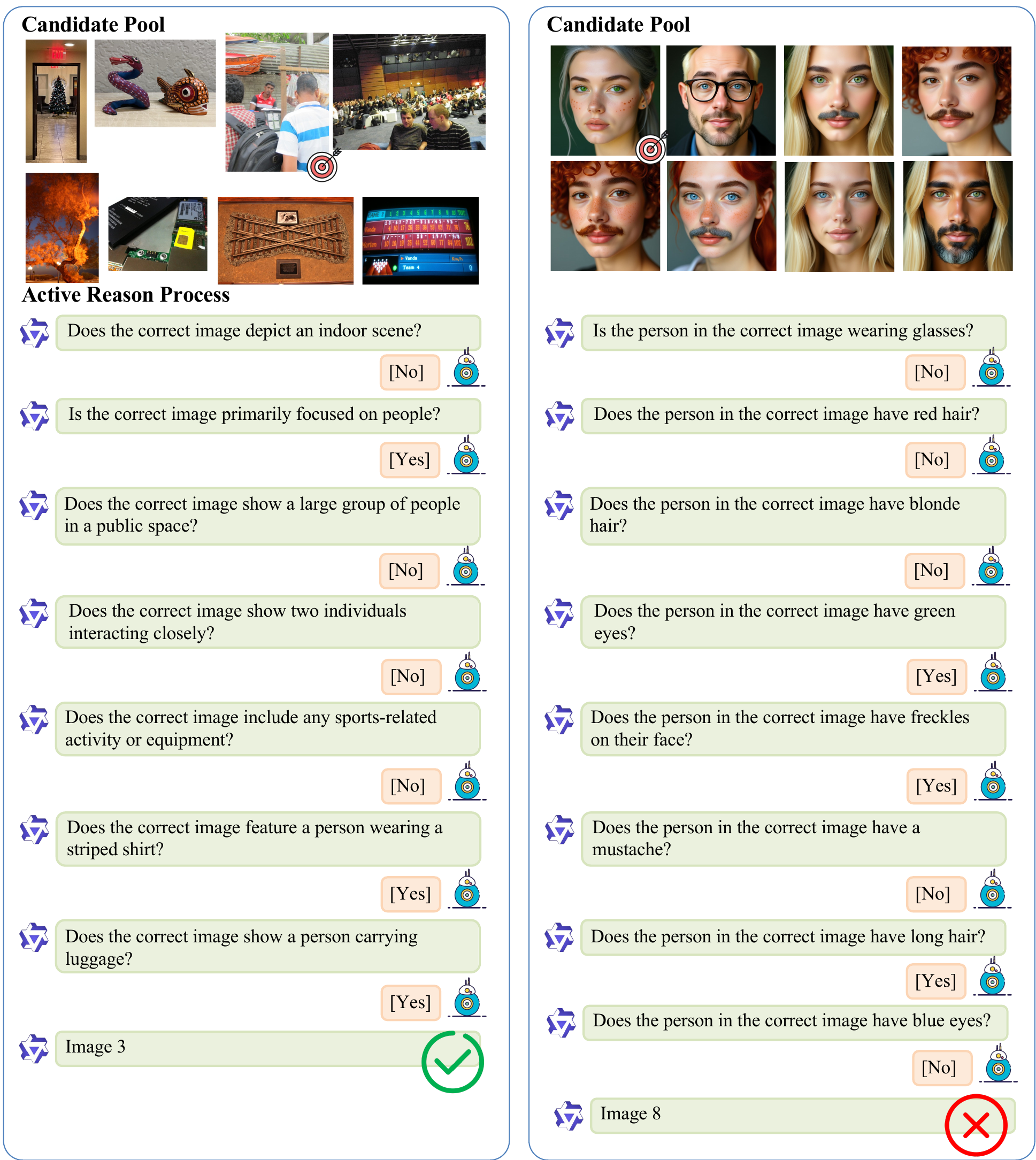}
    \caption{Case study: active reason. The left is Qwen2.5-VL-72B model, and the right is Qwen2.5-VL-32B model.}
    \label{app-fig:case1}
\end{figure*}

\begin{figure*}[t]
  \centering
  % 同时限制宽高，上不超过正文宽度与高度；保持纵横比
  \includegraphics[
    width=\textwidth,
    height=\dimexpr\textheight-\abovecaptionskip-\belowcaptionskip\relax,
    keepaspectratio
  ]{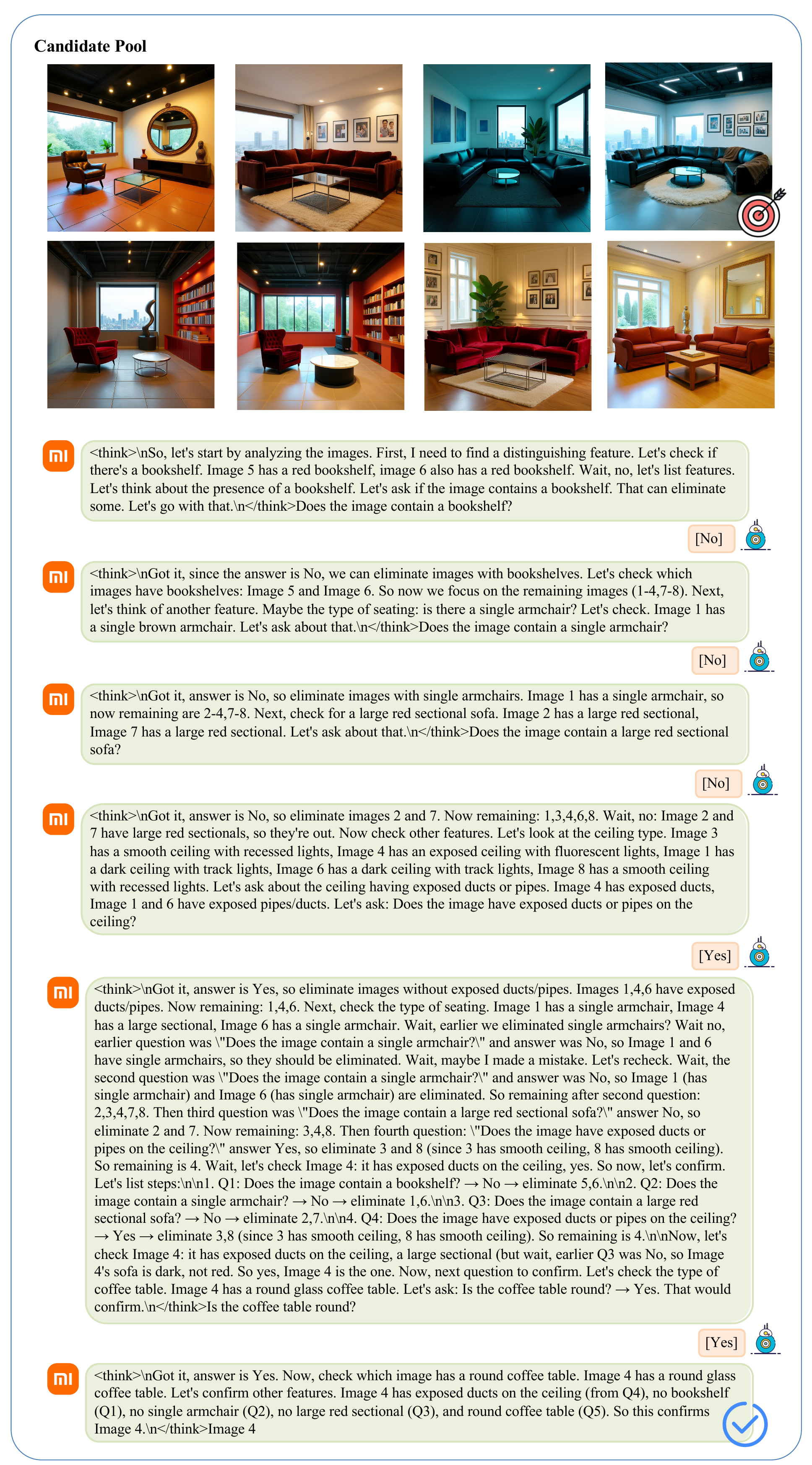}
  \caption{Case study: active reasoning with explicit thinking. The evaluated model is Mimo-VL model.}
  \label{app-fig:case2}
\end{figure*}

We provide three cases for illustration in Figure~\ref{app-fig:case1} and Figure~\ref{app-fig:case2}, including a good case, a bad case, and an explicit thinking case.

%% file: page/02Related_works.tex
\section{Related Works}
% \subsection{Multimodal Large Language Models}
% Multimodal large language models (MLLMs) have emerged as powerful systems, demonstrating strong capabilities in both perception and reasoning. Numerous methods have been proposed to further enhance their capabilities in these two areas, such as dynamic resolution, chain-of-thought reasoning, and thinking with images, achieving strong results on various benchmarks. However, existing benchmarks primarily evaluate passive reasoning and fail to assess model performance under conditions of missing or incomplete evidence. As a result, current models lack an accurate way to measure and ultimately improve their ability in active reasoning. To address this limitation, we propose GuessBench, a new dataset that explicitly targets active reasoning, providing both a benchmark for evaluation and a direction for improving future MLLMs.

% \subsection{Multimodal Benchmarks}
There have been numerous multimodal benchmarks for evaluating MLLMs across real-world scenarios~\cite{liu2024seeing,zhou2025micevalunveilingmultimodalchain}. For example, POPE~\cite{Li2023EvaluatingOH} targets object perception, HallusionBench~\cite{Guan_2024_CVPR} examines image–text hallucination, and dynamicME~\cite{Yang2024DynamicME} further explores the performance under generative images. To move beyond single-capability probes and toward broader utility, subsequent work has emphasized complementary axes. MME~\cite{fu2024mmecomprehensiveevaluationbenchmark} assesses knowledge utilization, and MathVista~\cite{lu2024mathvista} evaluates visual mathematics. Reasoning-oriented evaluation has become a central focus in recent work, including M$^3$CoT~\cite{chen-etal-2024-m3cot}, MMMU~\cite{yue2023mmmu}, and so on~\cite{zhang2024cmmmu,liu2023mmc}.
However, most benchmarks still present models with full information at inference time and thus primarily test passive reasoning. This setting diverges from real-world use, where models must reason actively under incomplete or ambiguous evidence. To bridge this gap, we propose GuessBench, the first benchmark that systematically evaluates the active reasoning capability of multimodal large language models, complementing and extending existing evaluations.

%% file: table/human_label.tex
\begin{table}[htbp!]
   \centering
   \small
   \begin{tabular}{lc}
     \toprule
     \textbf{Model} & \textbf{Reliability Rate} \\
     \midrule
     InternVL3-2B & 99.01\% \\
     InternVL3-8B & 98.71\% \\
     InternVL3-14B & 94.12\% \\
     InternVL3-38B & 98.34\% \\
     InternVL3-78B & 97.82\% \\
     \midrule
     Qwen2.5-VL-7B & 98.74\% \\
     Qwen2.5-VL-32B & 98.58\% \\
     Qwen2.5-VL-3B & 98.33\% \\
     Qwen2.5-VL-72B & 98.01\% \\
     \midrule
     InternVL3.5-2B & 100.00\% \\
     InternVL3.5-4B & 96.14\% \\
     InternVL3.5-8B & 98.25\% \\
     InternVL3.5-14B & 99.39\% \\
     InternVL3.5-38B & 98.15\% \\
     \midrule
     Kimi-VL-A3B & 98.07\% \\
     Kimi-VL-A3B-Thinking & 95.98\% \\
     Qwen3-VL-Instruct &98.28\% \\
     Qwen3-VL-Thinking & 98.41\%\\
     MiMo-VL & 99.04\% \\
     GPT-4o-mini & 98.16\% \\
     \bottomrule
   \end{tabular}
   \caption{The result of the evaluation procedure by human verification.}
   \label{tab:human_rate}
\end{table}

%% file: table/prompt/base_prompt.tex
\begin{table*}[t]
\centering
\small
\setlength{\tabcolsep}{8pt}
\renewcommand{\arraystretch}{1.15}

\begin{tabularx}{\textwidth}{X}
\rowcolor[HTML]{343434}
\multicolumn{1}{c}{\color{white}\textbf{Prompt}}\\
\toprule

\rowcolor[HTML]{F0F0F0}
\multicolumn{1}{c}{\textbf{Real images and perception-oriented images}}\\
You are presented with a series of images, from which one is the correct answer. Your task is to identify this correct image by asking a series of questions within 10 turns. You must use all available images to formulate your questions, and ultimately output the number corresponding to the correct image.\\
\textbf{Instructions}\\
1. \textbf{Objective:} Identify the correct image among the provided options by asking yes/no questions.\\
2. \textbf{Question Format:} Each question must be a single, clear, and concise \textbf{yes/no question}. Do not request additional hints.\\
3. \textbf{Iterative Process:} After each question, you will receive a ``[Yes]'' or ``[No]'' answer. Use this information, along with previous answers, to formulate your next question.\\
4. \textbf{Final Answer:} When you are confident you have identified the correct image, state its corresponding number as your final answer (e.g., ``Image 3'').\\
5. \textbf{Efficiency:} The goal is to identify the correct image with the fewest possible questions.\\
6. \textbf{Repeating the same question is strictly prohibited.} Each question must aim to gather new information and actively eliminate a significant number of remaining incorrect images. Questions should progress logically from broader characteristics to more specific details.\\
\textbf{Start}\\
Let's begin. Here are the images you need to consider:\\
\midrule
\rowcolor[HTML]{F0F0F0}
\multicolumn{1}{c}{\textbf{Knowledge-oriented images}}\\
You are presented with a series of images, from which one is the correct answer. Your task is to identify this correct image by asking a series of questions within 10 turns. You must use all available images to formulate your questions, and ultimately output the number corresponding to the correct image.\\
\textbf{Instructions}\\
1. \textbf{Objective:} Identify the correct image among the provided options by asking yes/no questions.\\
2. \textbf{Question Format:} Each question must be a single, clear, and concise \textbf{yes/no question}. Do not request additional hints.\\
3. \textbf{Iterative Process:} After each question, you will receive a ``[Yes]'' or ``[No]'' answer. Use this information, along with previous answers, to formulate your next question.\\
4. \textbf{Final Answer:} When you are confident you have identified the correct image, state its corresponding number as your final answer (e.g., ``Image 3'').\\
5. \textbf{Efficiency:} The goal is to identify the correct image with the fewest possible questions.\\
6. \textbf{Repeating the same question is strictly prohibited.} Each question must aim to gather new information and actively eliminate a significant number of remaining incorrect images. \\
7. \textbf{Questions should progress logically from broader characteristics to more specific details}, emphasizing the underlying knowledge, function, history, cultural significance, scientific principles, or relationships to other objects that differentiate them.\\
\textbf{Start}\\
Let's begin. Here are the images you need to consider:\\
\bottomrule
\end{tabularx}

\caption{Prompt for multimodal large language models.}
\label{app-tab:prompt_mllm}
\end{table*}

%% file: table/prompt/guess_agent.tex
\begin{table*}[t]
\centering
\small
\setlength{\tabcolsep}{8pt}
\renewcommand{\arraystretch}{1.15}

\begin{tabularx}{\textwidth}{X}
\rowcolor[HTML]{343434}
\multicolumn{1}{c}{\color{white}\textbf{Prompt for state detector}}\\
\toprule
You are a text classifier. Your task is to analyze the user's input and determine its type.\\
\textbf{Classification Rules:}\\
1.  \textbf{Output <QUERY>}: If the input is clearly a \textbf{question}. A question typically ends with a question mark (?) or is phrased in an interrogative manner.\\
2.  \textbf{Output <STOP>}: If the input is in the format of \textbf{"Image [number]"} (e.g., "Image 3", "Image 10").\\
3. \textbf{Output <INVALID>}: if the input cannot be answer by yes/no.\\
Here is the text:\\
\bottomrule
\end{tabularx}

\caption{Prompt for state detector.}
\label{app-tab:prompt_state}
\end{table*}

\begin{table*}[t]
\centering
\small
\setlength{\tabcolsep}{8pt}
\renewcommand{\arraystretch}{1.15}

\begin{tabularx}{\textwidth}{X}
\rowcolor[HTML]{343434}
\multicolumn{1}{c}{\color{white}\textbf{Prompt for evidence provided by LLM}}\\
\rowcolor[HTML]{F0F0F0} \multicolumn{1}{c}{\textbf{Prompt for evidence provided by LLM}}\\
You are the judge in an image selection game. Your role is to evaluate player questions against a set of given images, one of which has been secretly designated as the \textbf{correct image}. You must answer player questions truthfully and objectively based on the description of this chosen image.\\
\textbf{Judging Rules:}\\
1.  \textbf{Response Format:} You may only use four standard responses: \textbf{"[Yes]"}, \textbf{"[No]"} and \textbf{"[Invalid]"}. Provide no additional explanations or hints. Maintain absolute objectivity and do not reveal any specific information about the correct image beyond the yes/no answer.\\
        2.  \textbf{Question Evaluation:}\\
             Respond with \textbf{"[Yes]"} or \textbf{"[No]"} if the question can be answered based on the description of the \textbf{correct image}.\\
             Respond with \textbf{"[Invalid]"} if the question can not be answered based on the description of the \textbf{correct image}.\\
        \textbf{Judge's Responsibilities:}\\
         Maintain absolute neutrality; provide no additional hints.\\
         Strictly follow the above rules to ensure the fairness of the game process.\\
        \textbf{The description of the correct image is: }\\
        \textbf{The question is: }\\
\midrule
\rowcolor[HTML]{F0F0F0}
\multicolumn{1}{c}{\textbf{Prompt for evidence provided by MLLM}}\\
        You are the judge in an image selection game. Your role is to evaluate player questions against a set of given images, one of which has been secretly designated as the \textbf{correct image}. You must answer player questions truthfully and objectively based on the description of this chosen image.\\
        \textbf{Judging Rules:}\\
        1.  \textbf{Response Format:} You may only use four standard responses: \textbf{"[Yes]"}, \textbf{"[No]"} and \textbf{"[Invalid]"}. Provide no additional explanations or hints. Maintain absolute objectivity and do not reveal any specific information about the correct image beyond the yes/no answer.\\
        2.  \textbf{Question Evaluation:}\\
             Respond with \textbf{"[Yes]"} or \textbf{"[No]"} if the question can be answered based on the description of the \textbf{correct image}.\\
             Respond with \textbf{"[Invalid]"} if the question can not be answered based on the description of the \textbf{correct image}.\\
        \textbf{Judge's Responsibilities:}\\
         Maintain absolute neutrality; provide no additional hints.\\
         Strictly follow the above rules to ensure the fairness of the game process.\\
        \textbf{The correct image is: }\\
        \textbf{The question is: }\\
\bottomrule
\end{tabularx}

\caption{Prompt for evidence provided.}
\label{app-tab:answer}
\end{table*}

%% file: table/prompt/method_prompt.tex
\begin{table*}[t]
\centering
\small
\setlength{\tabcolsep}{8pt}
\renewcommand{\arraystretch}{1.15}

\begin{tabularx}{\textwidth}{X}
\rowcolor[HTML]{343434}
\multicolumn{1}{c}{\color{white}\textbf{Prompt}}\\
\toprule

\rowcolor[HTML]{F0F0F0} \multicolumn{1}{c}{\textbf{Real images and perception-oriented images}}\\
You are presented with a series of images, from which one is the correct answer. Your task is to identify this correct image by asking a series of questions within 10 turns. You must use all available images to formulate your questions, and ultimately output the number corresponding to the correct image.\\
\textbf{Instructions}\\
1. \textbf{Objective:} Identify the correct image among the provided options by asking yes/no questions.\\
2. \textbf{Question Format:} Each question must be a single, clear, and concise \textbf{yes/no question}. Do not request additional hints.\\
3. \textbf{Iterative Process:} After each question, you will receive a ``[Yes]'' or ``[No]'' answer. Use this information, along with previous answers, to formulate your next question.\\
4. \textbf{Final Answer:} When you are confident you have identified the correct image, state its corresponding number as your final answer (e.g., ``Image 3'').\\
5. \textbf{Efficiency:} The goal is to identify the correct image with the fewest possible questions.\\
6. \textbf{Repeating the same question is strictly prohibited.} Each question must aim to gather new information and actively eliminate a significant number of remaining incorrect images. Questions should progress logically from broader characteristics to more specific details.\\
\textbf{Intelligent Guessing Strategy}\\
        1.  \textbf{Chained Reasoning, Step by Step}: Before each question, construct a rigorous chain of reasoning based on your current clues. Clearly define all current possibilities and precisely pinpoint the "decisive question" that will most effectively eliminate distractions and rapidly lead you closer to the truth.\\
        2.  \textbf{Feature Filtering, Efficient Focus}: Prioritize asking about key features that possess strong differentiating power. Your goal is to swiftly eliminate a large number of ineligible cards, efficiently narrowing your focus to a few high-potential options.\\
        3.  \textbf{Know When to Stop, Avoid Redundancy}: Once you're confident about the target card, make your judgment decisively. Avoid unnecessary additional questions; these not only waste opportunities but could also negatively impact your score.
        Let's begin. \\
\textbf{Start}\\
Let's begin. Here are the images you need to consider:\\
\midrule
\rowcolor[HTML]{F0F0F0} \multicolumn{1}{c}{\textbf{Knowledge-oriented images}}\\
You are presented with a series of images, from which one is the correct answer. Your task is to identify this correct image by asking a series of questions within 10 turns. You must use all available images to formulate your questions, and ultimately output the number corresponding to the correct image.\\
\textbf{Instructions}\\
1. \textbf{Objective:} Identify the correct image among the provided options by asking yes/no questions.\\
2. \textbf{Question Format:} Each question must be a single, clear, and concise \textbf{yes/no question}. Do not request additional hints.\\
3. \textbf{Iterative Process:} After each question, you will receive a ``[Yes]'' or ``[No]'' answer. Use this information, along with previous answers, to formulate your next question.\\
4. \textbf{Final Answer:} When you are confident you have identified the correct image, state its corresponding number as your final answer (e.g., ``Image 3'').\\
5. \textbf{Efficiency:} The goal is to identify the correct image with the fewest possible questions.\\
6. \textbf{Repeating the same question is strictly prohibited.} Each question must aim to gather new information and actively eliminate a significant number of remaining incorrect images. \\
7. \textbf{Questions should progress logically from broader characteristics to more specific details}, emphasizing the underlying knowledge, function, history, cultural significance, scientific principles, or relationships to other objects that differentiate them.\\
\textbf{Intelligent Guessing Strategy}\\
        1.  \textbf{Chained Reasoning, Step by Step}: Before each question, construct a rigorous chain of reasoning based on your current clues. Clearly define all current possibilities and precisely pinpoint the "decisive question" that will most effectively eliminate distractions and rapidly lead you closer to the truth.\\
        2.  \textbf{Feature Filtering, Efficient Focus}: Prioritize asking about key features that possess strong differentiating power. Your goal is to swiftly eliminate a large number of ineligible cards, efficiently narrowing your focus to a few high-potential options.\\
        3.  \textbf{Know When to Stop, Avoid Redundancy}: Once you're confident about the target card, make your judgment decisively. Avoid unnecessary additional questions; these not only waste opportunities but could also negatively impact your score.
        Let's begin. \\
\textbf{Start}\\
Let's begin. Here are the images you need to consider:\\
\bottomrule
\end{tabularx}

\caption{Prompt for chain-of-thought.}
\label{app-tab:prompt_cot}
\end{table*}

\begin{table*}[t]
\centering
\small
\setlength{\tabcolsep}{8pt}
\renewcommand{\arraystretch}{1.15}

\begin{tabularx}{\textwidth}{X}
\rowcolor[HTML]{343434}
\multicolumn{1}{c}{\color{white}\textbf{Prompt}}\\
\toprule

\rowcolor[HTML]{F0F0F0} \multicolumn{1}{c}{\textbf{Knowledge background prompt}}\\
\textbf{real world}:Subject Matter, Camera Angle, Shot Type and Composition, Lighting and Atmosphere, Color and Specific Details.\\
\textbf{face}:  hair style and colore, eyes colore and shape, eyebrows shape and color, nose shape, lips thickness and color, beard type and color, skin tone and marking, face shape, glasses.\\
\textbf{animal}:  Mammals, Birds, Reptiles, Fish, Insects and Arachnids, Farm Animals, Extinct.\\
\textbf{city}:  Time of Day, Weather, Street Type, Buildings, Vehicles, People Activity, Street Elements.\\
\textbf{kitchen}:  Style, Cabinets color and style, Countertop, Appliances stove and fridge, Flooring, Backsplash, Island Presence and feature, Lighting, Decor.\\
\textbf{livingroom}: Style, sofa type and material, coffee table material and shape, wall decor, lighting, color scheme, flooring, window view, notable accessory.\\
\textbf{nature}:  Terrain, Time of Day, Weather Condition, Vegetation, Water Feature, Wildlife Element, Light Quality, Overall Atmosphere.\\
\textbf{profession}:  professions, actions, environments, styles moods.\\
\textbf{traffic}:  traffic sigh.\\
\textbf{vehicle}:  vehicle types, colors, materials details, conditions, eras styles, environments, function.\\
\midrule
\rowcolor[HTML]{F0F0F0} \multicolumn{1}{c}{\textbf{Real images and perception-oriented images}}\\
You are presented with a series of images, from which one is the correct answer. Your task is to identify this correct image by asking a series of questions within 10 turns. You must use all available images to formulate your questions, and ultimately output the number corresponding to the correct image.\\
\textbf{Instructions}\\
1. \textbf{Objective:} Identify the correct image among the provided options by asking yes/no questions.\\
2. \textbf{Question Format:} Each question must be a single, clear, and concise \textbf{yes/no question}. Do not request additional hints.\\
3. \textbf{Iterative Process:} After each question, you will receive a ``[Yes]'' or ``[No]'' answer. Use this information, along with previous answers, to formulate your next question.\\
4. \textbf{Final Answer:} When you are confident you have identified the correct image, state its corresponding number as your final answer (e.g., ``Image 3'').\\
5. \textbf{Efficiency:} The goal is to identify the correct image with the fewest possible questions.\\
6. \textbf{Repeating the same question is strictly prohibited.} Each question must aim to gather new information and actively eliminate a significant number of remaining incorrect images. Questions should progress logically from broader characteristics to more specific details.\\
\textbf{Domain Background}\\
Here's the domain-specific knowledge you can use to guide your guesses as a player: \textcolor{blue}{knowledge background} and \textcolor{blue}{knowledge background prompt}\\
\textbf{Start}\\
Let's begin. Here are the images you need to consider:\\
\midrule
\rowcolor[HTML]{F0F0F0} \multicolumn{1}{c}{\textbf{Knowledge-oriented images}}\\
You are presented with a series of images, from which one is the correct answer. Your task is to identify this correct image by asking a series of questions within 10 turns. You must use all available images to formulate your questions, and ultimately output the number corresponding to the correct image.\\
\textbf{Instructions}\\
1. \textbf{Objective:} Identify the correct image among the provided options by asking yes/no questions.\\
2. \textbf{Question Format:} Each question must be a single, clear, and concise \textbf{yes/no question}. Do not request additional hints.\\
3. \textbf{Iterative Process:} After each question, you will receive a ``[Yes]'' or ``[No]'' answer. Use this information, along with previous answers, to formulate your next question.\\
4. \textbf{Final Answer:} When you are confident you have identified the correct image, state its corresponding number as your final answer (e.g., ``Image 3'').\\
5. \textbf{Efficiency:} The goal is to identify the correct image with the fewest possible questions.\\
6. \textbf{Repeating the same question is strictly prohibited.} Each question must aim to gather new information and actively eliminate a significant number of remaining incorrect images. \\
7. \textbf{Questions should progress logically from broader characteristics to more specific details}, emphasizing the underlying knowledge, function, history, cultural significance, scientific principles, or relationships to other objects that differentiate them.\\
\textbf{Domain Background}\\
Here's the domain-specific knowledge you can use to guide your guesses as a player: \textcolor{blue}{knowledge background} and \textcolor{blue}{knowledge background prompt}\\
        
\textbf{Start}\\
Let's begin. Here are the images you need to consider:\\
\bottomrule
\end{tabularx}

\caption{Prompt for reason prompt. The \textcolor{blue}{text} denotes denotes the input information.}
\label{app-tab:prompt_reason_prompt}
\end{table*}

\begin{table*}[t]
\centering
\small
\setlength{\tabcolsep}{8pt}
\renewcommand{\arraystretch}{1.15}

\begin{tabularx}{\textwidth}{X}
\rowcolor[HTML]{343434}
\multicolumn{1}{c}{\color{white}\textbf{Prompt for RAG model}}\\
\toprule
\textbf{Task:}\\
            Answer this question based on the image.\\
\textbf{Note:}\\
            - Your output must be exclusively and only [Yes] or [No].\\
\bottomrule
\end{tabularx}

\caption{Prompt for RAG models.}
\label{app-tab:prompt_rag}
\end{table*}

\begin{table*}[t]
\centering
\small
\setlength{\tabcolsep}{8pt}
\renewcommand{\arraystretch}{1.15}

\begin{tabularx}{\textwidth}{X}
\rowcolor[HTML]{343434}
\multicolumn{1}{c}{\color{white}\textbf{Prompt for ReAct}}\\
\toprule
Based on the previous questions and the [Yes/No] answer you received, consider which images have been eliminated and what information you still need to narrow down the possibilities. Specifically, detail: \\
    1.  \textbf{Which images have been eliminated} based on all previous questions and their corresponding [Yes/No] answers.\\
    2.  \textbf{Why these images were eliminated} (i.e., which specific characteristics or features contradicted the answers).\\
    3.  \textbf{What characteristics or features are you now focusing on} for the remaining possible images.\\
    4.  \textbf{What is your strategic approach for the next question?} Are you trying to broadly eliminate more images, or narrow down specific details of a few remaining options?\\
    The history is: \\
\bottomrule
\end{tabularx}

\caption{Prompt for ReAct.}
\label{app-tab:prompt_react}
\end{table*}